%% file: main.tex
\documentclass[10pt,twocolumn,letterpaper]{article}

\usepackage{iccv}              % To produce the CAMERA-READY version
\pdfoutput=1

\pdfoutput=1
\usepackage{longtable}
\usepackage{tabularx}
\usepackage{array}
\usepackage{booktabs} % Optional: Improves table formatting
\usepackage[utf8]{inputenc}
\usepackage{lscape}
\usepackage{microtype} % Better text justification and hyphenation
\usepackage{float}
\usepackage{tabularx}

\def\eg{\emph{e.g., }}

\def\etal{\emph{et al. }}

\def\name{\emph{PVChat}}

\newcommand{\bfsection}[1]{\vspace*{0.0mm}\noindent\textbf{#1.}}

\usepackage{xcolor}
\usepackage{marvosym}
\definecolor{iccvblue}{rgb}{0.21,0.49,0.74}
\usepackage[pagebackref,breaklinks,colorlinks,allcolors=iccvblue]{hyperref}

\def\paperID{5019} % *** Enter the Paper ID here
\def\confName{ICCV}
\def\confYear{2025}

\title{\name{}: Personalized Video Chat with One-Shot Learning}

\author{
    Yufei Shi$^{1,5\dagger}$,
    Weilong Yan$^{2\dagger}$,
    Gang Xu$^{4}$,
    Yumeng Li$^{3}$,
    Yucheng Chen$^{1,5}$,\\
    Zhenxi Li$^{1,5}$,
    Fei Richard Yu$^{4}$,
    Ming Li$^{4}$\textsuperscript{\Letter},
    Si Yong Yeo$^{1,5}$\textsuperscript{\Letter}
    \\[10pt]
    $^{1}$MedVisAI Lab \quad
    $^{2}$National University of Singapore \quad
    $^{3}$Nankai University\\
    $^{4}$Guangdong Laboratory of Artificial Intelligence and Digital Economy (SZ)\\
    $^{5}$Lee Kong Chian School of Medicine, Nanyang Technological University\\[6pt]
    {\tt\small $^\dagger$Equal contribution. Email: \{yufei005@e.ntu.edu.sg, yanweilong@u.nus.edu\}}
}

\begin{document}
\maketitle

% Place the figure right after the title

\input{0_abstract}    
\begin{figure}[!t]  % !ht for "here" or "top" with high priority
    \centering
    \includegraphics[width=\columnwidth]{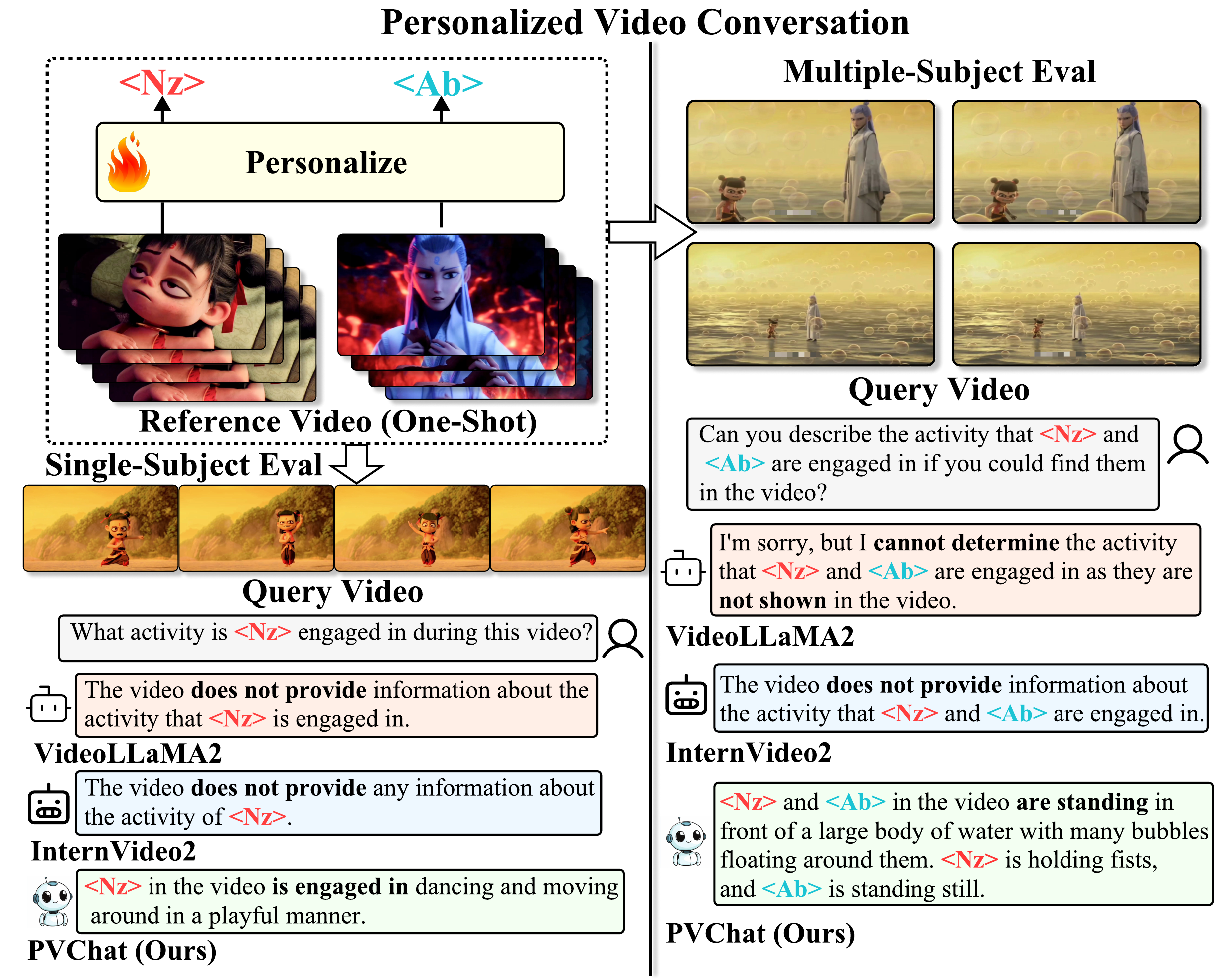}
    \caption{Examples of PVChat's ability with one-shot learning (e.g., \textless Nz\textgreater and \textless Ab\textgreater). PVChat can answer questions about the personalized information correctly while other models \cite{videollama, internvideo2} fail.}
    \label{fig:your_reference_label}
\end{figure}
\input{1_intro}

\input{2_related_work}
\input{3_method}
\input{4_experiment}

\input{5_conclusion}

    \small
    \bibliographystyle{ieeenat_fullname}
    \bibliography{main}
% Add a clear page break before supplementary material
\clearpage

% Reset section counters for supplementary material
\setcounter{section}{0}
\setcounter{figure}{0}
\setcounter{table}{0}
\setcounter{equation}{0}
\renewcommand{\thesection}{S\arabic{section}}
\renewcommand{\thefigure}{\arabic{figure}}  % Supplementary Figure
\renewcommand{\thetable}{\arabic{table}}    % Supplementary Table
\renewcommand{\theequation}{\arabic{equation}}  % Supplementary Equation

% Supplementary Material Title
% 使用单栏模式来显示标题
% 标题跨越双栏但不分页
\twocolumn[{
\begin{@twocolumnfalse}
    \begin{center}
        \vspace{1cm}  % 可以调整这个值来控制标题上方的空间
        \Large\textbf{Supplementary Material:\\
        \name{}: Personalized Video Chat with One-Shot Learning}
        \vspace{0.5cm}  % 可以调整这个值来控制标题下方的空间
    \end{center}
\end{@twocolumnfalse}
}]

\input{supp}  % Assuming your supplementary content is in a file called supp.tex in the sec directory

\end{document}

%% file: 0_abstract.tex
\begin{abstract}
Video large language models (ViLLMs) excel in general video understanding, e.g., recognizing activities like talking and eating, but struggle with identity-aware comprehension, such as ``Wilson is receiving chemotherapy" or ``Tom is discussing with Sarah", limiting their applicability in smart healthcare and smart home environments.
To address this limitation, we propose a one-shot learning framework \name{}, the first personalized ViLLM that enables subject-aware question answering (QA) from \emph{a single video for each subject}. Our approach optimizes a Mixture-of-Heads (MoH) enhanced ViLLM on a synthetically augmented video-QA dataset, leveraging a progressive image-to-video learning strategy. Specifically, we introduce an automated augmentation pipeline that synthesizes identity-preserving positive samples and retrieves hard negatives from existing video corpora, generating a diverse training dataset with four QA types: existence, appearance, action, and location inquiries.
To enhance subject-specific learning, we propose a ReLU Routing MoH attention mechanism, alongside two novel objectives: (1) Smooth Proximity Regularization for progressive learning through exponential distance scaling and (2) Head Activation Enhancement for balanced attention routing. Finally, we adopt a two-stage training strategy, transitioning from image pre-training to video fine-tuning, enabling a gradual learning process from static attributes to dynamic representations.
We evaluate \name{} on diverse datasets covering medical scenarios, TV series, anime, and real-world footage, demonstrating its superiority in personalized feature understanding after learning from a single video, compared to state-of-the-art ViLLMs. Code is at \href{https://github.com/DavidYan2001/PVChat}{PVChat}.
\end{abstract}

%% file: 1_intro.tex
\section{Introduction}
Recent advancements in Video Large Language Models (ViLLMs) have showcased impressive capabilities across a wide range of tasks, \eg video captioning, temporal action localization, and video question answering \cite{videollama,internvideo2,videochatgpt,chatvideo}. These models, trained on vast datasets, exhibit broad knowledge across multiple domains, such as medical assistance \cite{medllavavideo} and autonomous driving \cite{drivegpt4, Yan_2025_CVPR}, significantly extending the functional scope of ViLLMs.

Despite these achievements, current ViLLMs are largely confined to general-purpose content understanding and struggle in personalized scenarios that require distinguishing specific individuals. For instance, when analyzing a video featuring a person named ``Alex'', even state-of-the-art models fail to consistently recognize ``Alex'' in novel contexts, such as identifying him from others or describing his specific actions. This limitation arises because existing training data predominantly focus on general knowledge rather than learning identity-specific information about an individual. As a result, ViLLMs are ill-equipped for real-world applications that demand personalized comprehension, \eg smart home \cite{zhong2023viotgpt}, intelligent healthcare \cite{llavamed} and human-computer interaction \cite{HCI}.

% To enhance the ability of LLMs in capturing personalized visual clues, some recent research has focused on image understanding \cite{yollava,personalized,personalized-large}. For example, Yo-LLaVA \cite{yollava} introduces learnable token prompts to equip models with individualized understanding capabilities. Other studies \cite{personalized,personalized-large} construct specialized training datasets for personalized image understanding through expert-curated annotations. While these methods have shown some potentials, they remain limited to image-based understanding and lack the ability to capture dynamic visual clues from videos which contain much richer personalized information. This gap underscores the need for dedicated solutions that facilitate identity-aware comprehension in video-based contexts.
To enhance the ability of LLMs in capturing personalized cues, recent research has focused on personalized image understanding \cite{yollava,personalized,personalized-large}. For example, Yo-LLaVA \cite{yollava} acquires the capability to encode a customized subject into a collection of latent tokens by utilizing a small number of sample images representing the subject. Other studies \cite{personalized,personalized-large} construct specialized training datasets curated with expert annotations to enable personalized image recognition. While these methods have demonstrated promising results, they are inherently limited to static image-based understanding and cannot model dynamic, temporally evolving visual cues present in videos. Given that videos contain significantly richer personalized information, including motion patterns, interaction dynamics, and contextual dependencies, there is a pressing need for solutions that enable identity-aware comprehension in video-based contexts.

% To address these limitations, we introduce \name{}, the first personalized video understanding LLM capable of comprehending individual characteristics and supporting subject-specific question-answering learning with only a single reference video.

% We established a comprehensive dataset generation pipeline leveraging off-the-shelf tools. We selected high-definition facial videos from CelebV-HQ \cite{celebv} and extracted closely matching negative samples from the large-scale facial dataset Laion-Face-5B \cite{laionface5b}. Through tools such as ConsisID \cite{consisid}, we generated identity-consistent positive sample videos, enabling us to expand a single input video into 81 videos with corresponding 1,455 question-answer pairs, thereby achieving one-shot learning capabilities.

% We developed the ReMoH (RElu Routing Mixture-of-Heads Attention) module and designed two optimization objectives: the SP (Smooth Proximity) Loss that enables gradual distance-based learning through exponential scaling, and the AE (Activation Enhancement) Loss that balances the utilization of shared and routed attention heads. These innovations address the issues of gradient explosion and inactive routed attention heads in multi-head attention mechanisms. Our training methodology follows a 2-stage approach to overcome limited data challenges, progressing from image-summary to video-QA. This sequential process enables the model to first recognize individual appearances before understanding specific actions and multi-person interactions.

To address these limitations, we propose \name{}, the first personalized video understanding LLM capable of learning individual characteristics and supporting subject-specific question answering from a single reference video. 
% To enable effective one-shot learning, we propose a systematic data augmentation pipeline that constructs high-quality personalized training samples using a suite of off-the-shelf tools. The pipeline generates identity-preserving positive samples that retain the subject from the reference video while incorporating visually similar yet distinct negative samples to enhance model discrimination. It begins with facial attribute extraction and demographic categorization to ensure identity consistency. We leverage ConsisID \cite{consisid} and PhotoMaker \cite{photomaker} to synthesize high-fidelity videos across diverse scenarios while maintaining identity integrity. To further improve robustness against visually similar distractors, we introduce hard negative samples by retrieving similar faces from Laion-Face-5B \cite{laionface5b} and CelebV-HQ \cite{celebv}, followed by video synthesis for these identities. Finally, we construct question-answer pairs covering four key categories—person existence, appearance, action, and location—using InternVideo2 \cite{internvideo2} and ChatGPT-4o to ensure linguistic coherence and personalization. This comprehensive pipeline enhances model generalization and adaptability for personalized video-based understanding.
To enable one-shot learning, we propose a data augmentation pipeline that synthesizes high-quality personalized samples. It generates identity-preserving positives while incorporating visually similar negatives for enhanced discrimination. The process begins with facial attribute extraction and demographic categorization, followed by high-fidelity video synthesis using ConsisID \cite{consisid} and PhotoMaker \cite{photomaker}. Hard negatives are introduced by retrieving similar faces from Laion-Face-5B \cite{laionface5b} and CelebV-HQ \cite{celebv} with corresponding video synthesis. Finally, question-answer pairs covering four key categories—subject existence, appearance, action, and location—are generated via InternVideo2 \cite{internvideo2} and ChatGPT-4o to ensure linguistic coherence and personalization.

% To effectively learn various subject-specific features, we introduce a ReLU Routing Mixture-of-Heads (ReMoH) attention mechanism that employs ReLu activation to dynamically select the attention heads replacing the conventional softmax and top-k allocation strategy that is trained in a discontinuous, non-differentiable way, limiting their performance and scalability. Additionally, we propose two novel optimization objectives: (1) Smooth Proximity (SP) Loss, which facilitates progressive learning via exponential distance scaling, and (2) Activation Enhancement (AE) Loss, which balances shared and routed attention heads, mitigating issues such as gradient explosion and inactive heads in multi-head attention mechanisms.
% To enhance subject-specific feature learning, we propose the ReLU Routing Mixture-of-Heads (ReMoH) attention mechanism, which replaces conventional softmax-based and top-k head selection with a ReLU-driven dynamic routing strategy. Unlike prior methods that rely on discontinuous, non-differentiable selection, ReMoH enables smooth and scalable training. Additionally, we introduce two novel optimization objectives, \ie Smooth Proximity Regularization (SPR), which facilitates progressive learning via exponential distance scaling, and Activation Enhancement (AE) loss, which balances shared and routed attention heads. These two objectives effectively mitigate gradient explosion and inactive heads in attention learning.
To improve subject-specific feature learning, we introduce a ReLU Routing Mixture-of-Heads (ReMoH) attention mechanism, which replaces conventional softmax-based and top-k head selection with a ReLU-driven dynamic routing strategy, enabling smooth and scalable training. Additionally, we propose two novel optimization objectives: Smooth Proximity Regularization, which promotes progressive learning via exponential distance scaling, and Head Activation Enhancement, which balances shared and routed attention heads, effectively mitigating gradient explosion and inactive heads in multi-head attention mechanisms. 
% To effectively optimize our ReMoH inspired \name{} with limited data, we adopt a two-stage training strategy from static image learning to video temporal modelling. The first stage is to training our model with images and a simple summary task, while the second stage is to fine-tune the model with video QA tasks. This progressive learning strategy allows the model to first learn static identity attributes before advancing to dynamic action recognition and multi-person interactions.
To optimize our ReMoH-inspired \name{} under limited data conditions, we adopt a two-stage training strategy transitioning from static image learning to video temporal modeling. In the first stage, the model is trained on images with a simple summary task to capture static identity attributes. In the second stage, it is fine-tuned on video QA tasks to develop dynamic action recognition and multi-person interaction capabilities. This progressive learning framework enables a structured transition from static representations to complex spatiotemporal reasoning.
We verify \name{} across a diverse range of personalized scenarios, including healthcare, TV series, anime, and real-world scenes, requiring the recognition of one, two, or three individuals. Experimental results demonstrate that \name{} achieves state-of-the-art performance in individual information comprehension and identity-aware reasoning from a single reference video.

In summary, our main contributions of this work are listed as follows:
% \begin{itemize}
% \item We present PVChat, the first personalized video-language model that enhances LLMs with personalization capabilities using just a single video of a subject. This advancement enables personalized video understanding and user-specific question answering, establishing a strong foundation for future personalized video comprehension.
% \item We designed the ReMoH module to enhance the extraction and selection of individual characteristics in videos, complemented by Smooth Proximity Loss and Activation Enhancement Loss to improve training stability and feature extraction capabilities.
% \item We propose a systematic video augmentation and QA generation pipeline based on single-video inputs, and employ a two-stage approach to progressively learn subjects' appearances and actions. We constructed a dataset containing 304 original video examples with over 30,000 QA pairs for fine-tuning, which will be publicly released.
% \end{itemize}
\begin{itemize}
\item We introduce \name{}, the first ViLLM that extends LLMs with personalization capability from a single reference video. This enables personalized video understanding and user-specific question-answering with only one-shot learning, laying a strong foundation for future individual video comprehension.

\item We develop a systematic video augmentation and QA generation pipeline from a single reference video. The key is to exploit off-the-shelf toolboxes to generate identity-preserving positives and synthesize hard negatives for confusingly similar faces. To support the research, we construct a diverse dataset containing $6$ individual scenarios, $304$ original videos, $2{,}304$ extended generated videos with over $30{,}000$ QA pairs, which will be publicly released to facilitate future research.

\item We design the ReLU Routing Mixture-of-Heads module to enhance the effective extraction of individual-specific characteristics from videos. Additionally, we introduce Smooth Proximity Regularization and Head Activation Enhancement to improve training stability and attention head activation.
\end{itemize}

%% file: 2_related_work.tex
\section{Related Works}
\subsection{Image Large Language Models}
The remarkable language comprehension and reasoning capabilities of Large Language Models (LLMs)~\cite{achiam2023gpt,claude2024family,team2023gemini,touvron2023llama} have spurred significant interest in extending them into multimodal understanding. Pioneering works demonstrate different approaches for vision-language alignment: Flamingo~\cite{touvron2023llama} processes interleaved image-text inputs through cross-attention layers for diverse multimodal tasks, while BLIP-2~\cite{li2023blip} employs a Querying Transformer (Q-Former) to bridge frozen visual encoders with LLMs. Simpler architecture like LLaVA~\cite{liu2023visual} achieves effective feature alignment using lightweight MLP projectors. Recent research systematically explores critical components of Multimodal LLMs (MLLMs), including dynamic high-resolution processing strategies~\cite{chen2024far,liu2024llavanext}, instruction-tuning methodologies~\cite{li2024mvbench,liu2024points}, and optimal visual encoder selection~\cite{tong2025cambrian, wei2024vary}. Comprehensive design space analyses from MM1~\cite{mckinzie2024mm1} and Idefics2~\cite{laurenccon2025matters} further establish foundational principles for MLLM development.

\subsection{Video Large Language Models}
With progress in image-based multimodal language models (MLLMs), video understanding research has garnered great attention. Video-ChatGPT~\cite{maaz2023video} and Vally~\cite{luo2023valley} aggregate temporal features into compact tokens, while Video-LLaVA~\cite{lin2023video} unifies image-video representations through aligned projections before LLM integration. Video-Teller~\cite{liu2023video} underscores modality alignment in pre-training objectives, whereas PLLaVA~\cite{xu2024pllava} examines feature pooling impacts on video Question-Answering tasks. For extended sequences, LLaMA-VID~\cite{li2024llama} encodes frames with dual tokens to balance efficiency and information retention, while MovieChat~\cite{song2024moviechat} employs memory optimization for processing ultra-long videos (\textgreater 10K frames). Weng~\etal\cite{weng2024longvlm} enhance local segment features by injecting global semantics through hierarchical fusion mechanisms. These approaches collectively address spatiotemporal modeling challenges across varied input scales. However, these models are still incapable of understanding videos with personalization.

\subsection{Personalized Large Language Models}
Personalization in generative models exhibits distinct implementations across domains. For image generation, existing approaches focus on pixel-level subject fidelity through either concept token optimization~\cite{gal2022image, kumari2023multi} or model parameter adaptation~\cite{ruiz2023dreambooth}. In natural language processing contexts, personalization typically involves shaping LLMs' behavioral patterns (\eg, conversational style) via prompt engineering or metadata-driven retrieval mechanisms~\cite{liu2020you, zhang2018personalizing, peng2022godel}. Recent personalized modeling approaches employ parameter-efficient adaptation through few-shot learning. MyVLM~\cite{alaluf2024myvlm} implements a Q-former-based architecture with trainable projection heads for concept extraction and visual feature augmentation, whereas Yo'LLaVA~\cite{nguyen2025yo} extends the LLaVA framework by integrating novel concepts as specialized tokens within the LLM's embedding space. To the best of our knowledge, our model is the first to support videos input while others only accept images as inputs.

\label{sec:relatedwork}

%% file: 3_method.tex
\section{\name{}}
\label{sec:method}
\subsection{Overview}
To achieve video personalization understanding, we first establish a promising and systematic data augmentation pipeline in \ref{data production pipeline} to deal with the severe lack of diverse video data for individuals. To enhance the subject-specific learning, we propose a ReLU Routing MoH Attention mechanism in \ref{ReMoH}, alongside two novel objectives: (1) Smooth Proximity Regularization (SPR) for progressive learning through exponential distance scaling; (2) Head Activation Enhancement (HAE) for balanced attention routing. Finally, a two-stage training strategy is adopted in \ref{training pipeline} to progressively learn from static features to dynamic features of the target.

\subsection{Data Collection}
\label{data production pipeline}
\begin{figure*}[!t]  % !t表示优先尝试放在页面顶部
    \centering
    \includegraphics[width=\textwidth]{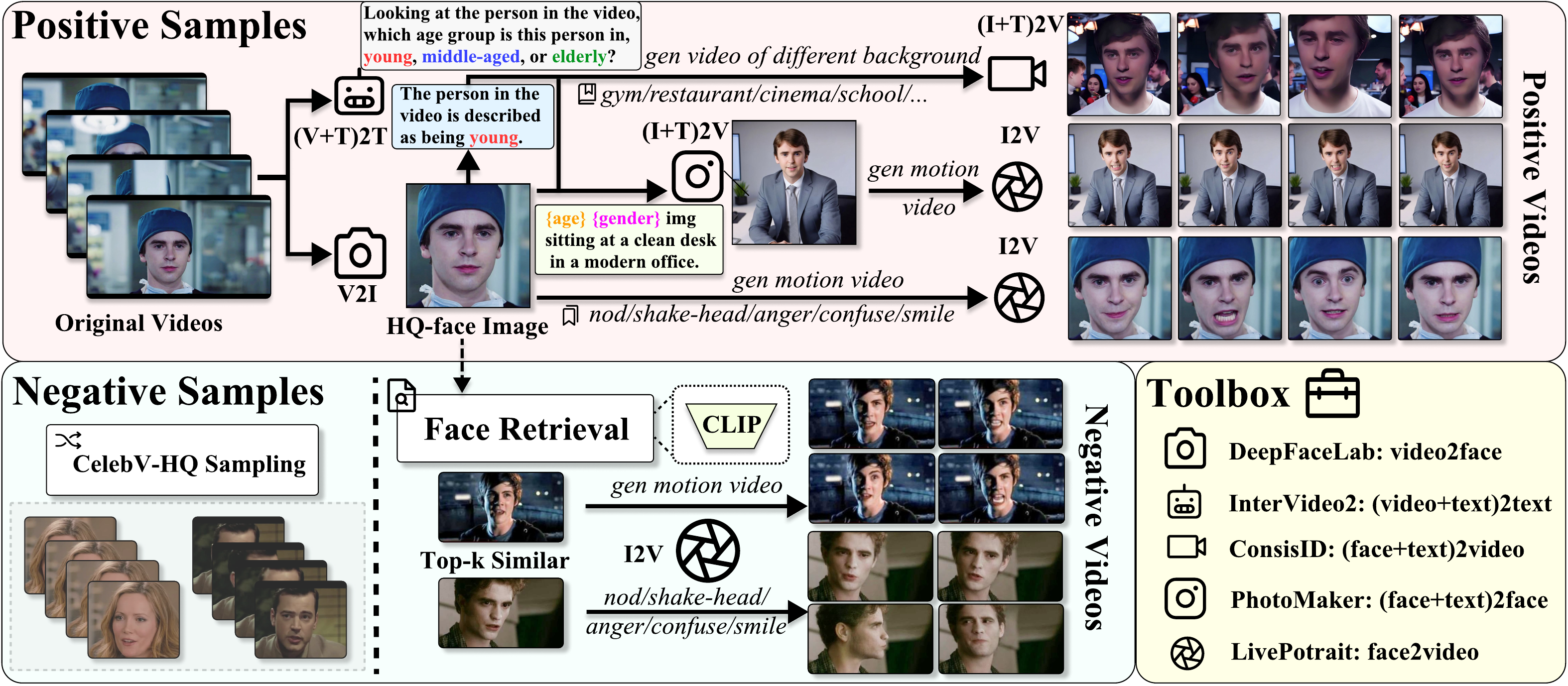}
    \caption{The systematic data collection pipeline. For positive data collection, the original videos are processed by DeepFaceLab \cite{deepfacelab} for high-quality face and InterVideo2 \cite{internvideo2} for demographic characteristics, which boost identity preservation. ConsisID \cite{consisid} and LivePortrait \cite{liveportrait} with PhotoMaker \cite{photomaker} utilize the identity information to generate videos of various background or different motion/expression, respectively. For model's robust perception, hard negative samples are selected from either similar face retrieval to generate negative videos, or sampled from the CelebV-HQ dataset \cite{celebv}. These negative samples guarantee the model's accurate recognition of both identity and content.}
    \label{fig:data_gen_pipeline}
    \vspace{-0.2cm}
\end{figure*}
% 为了解决小数据量训练的问题，并实现one-shot的能力，我们会对实验数据建立了一套完整的扩充流程
% 首先我们会利用deepfacelab工具，对视频中人脸进行提取，如果有多个脸，我们会利用InceptionResnetV1提取人脸的embedding，然后利用DBSCAN来分类，将多个人脸分开
%然后我们会来计算眼睛纵横比 (EAR)来判断人眼睛是否睁开，如果任意一只眼睛的EAR低于0.25（阈值），认为眼睛是闭合的
%判断是否是正脸，计算图像清晰度，人脸关键点是否存在，4个指标来打分，然后筛选出质量最好的一张人脸图片HQ-face

%然后我们会利用internvideo模型去问这个视频得到视频中人物的age和gender，得到的年龄一共3种，young，middle-aged，elderly，得到的性别一共两种male和female
%'According to the video and just told me two words about gender from "male, female" and age from "young, middle-aged, elderly" about the protagonist of the video',这个是问的脚本

%然后我们会利用ConsisID模型，和我们预先写好的几百个数量的prompt，（这里会有一个图来展示）根据之前得到的年龄和性别得到10个在不同的场景下的描述，然后使用consisid模型结合我们得到的高质量人脸去合成10个视频

%然后我们会利用Photomaker模型，根据我们的高质量人脸去生成在其他场景下的高保真id-Consis的图片，并结合性别年龄得到对应的prompt（具体的放在了附录中）

%然后我们会利用HQ-face搜索Laion-face-5b中最相似的一些图片来当做很难分类的负样本图片

%然后我们会利用LivePortrait模型，按照我们预设的5个动作，点头yes，摇头no，疑问query，愤怒angry，微笑smile，让这里的hq-face，以及Photomaker生成的图片动起来，并且让难以分类的负样本图片动起来充当负样本
%并且会随机从celebv-hq数据集中随机选30个视频当做我们的负样本视频

%然后我们会通过internvideo2模型来询问这些模型，4类问题，是否存在<sks>（放到了附录表格中），穿着是什么，在哪里，在干什么（预先设计好的15个问题，附录x），然后回复得到关于the person 在做什么后，再通过chatgpt4o的api去找到关于general的人类的描述，换成这里我们需要的<person>，（放到了附录里，关于chatgpt4o问询的prompt）
%最终我们由1个视频扩充得到81个视频以及1455个QA pairs。

To address the challenge of limited training data and achieve one-shot learning, we propose a novel and comprehensive data augmentation pipeline as illustrated in Fig.\ref{fig:data_gen_pipeline}, which starts from two key principles: 1) enabling automated personalized data collection directly from a wide range of natural videos; 2) ensuring that the generated personalized data encompasses different scenes, actions, and contextual elements while preserving the identity information.

\bfsection{Identity-Preserving Positives Generation}
Maintaining the identity information is key to obtaining diverse data of a specific individual. To achieve that, our pipeline starts with facial extraction from original videos using DeepFaceLab \cite{deepfacelab}. 
When multiple faces are present, we carefully design a two-stage approach: 
first, we extract facial embeddings via FaceNet \cite{FaceNet}, then apply DBSCAN \cite{dbscan} clustering to differentiate between subjects.
After that, the faces are evaluated using several metrics, including Eye Aspect Ratio (EAR) to determine eye openness (with a threshold of 0.25), facial orientation, image clarity, and presence of facial landmarks. 
These metrics allow us to identify the highest-quality facial image (HQ-face) for subsequent steps.

To further enhance the stability of identity preservation, the characteristics of individuals in the videos are determined using InternVideo2 \cite{internvideo2}, which are categorized by gender (male/female) and age group (young/middle-aged/elderly) based on the video content and our designed prompt.
With these demographic characteristics, we can more accurately preserve the identity, and employ ConsisID \cite{consisid} with our extensive prompt library Fig. \ref{fig:bing} to generate a wide range of videos in different scenarios (e.g., gym, restaurant, school, etc) with identity consistency.
%
% This approach enables us to synthesize 10 unique videos depicting the subject in various scenarios (e.g., gym, restaurant, school, etc) while maintaining identity consistency. 
%
Although videos from ConsisID \cite{consisid} contain rich content and we extract multiple types of identity information, their ID consistency is still poor, leading to a decline in data quality. 

With the aim of incorporating more ID consistent data, 
we utilize PhotoMaker \cite{photomaker} with the identified characteristics to generate additional high-fidelity images of the subject across different environments, and adopt LivePortrait \cite{liveportrait} to animate these images to generate high-identity-consistency videos with a facial motion and expression library (including nodding, head-shaking, questioning, anger, and smiling). 
Meanwhile, we also directly input the HQ-face image to LivePortrait \cite{liveportrait}, generating the motion video.
These videos exhibit strong ID consistency but include simple contents, complementing the data from ConsisID \cite{consisid}.
% with prompts tailored to the subject's identified demographic characteristics.

\bfsection{Hard Negatives Retrieval}
Avoiding the misidentification of personalized subjects is crucial for personalized video chatting.
As observed in previous VLM research \cite{yollava} and our video chat experiments, training with only positive data can lead to the model losing its ability to recognize personalized subjects.
Considering these factors, we introduce challenging negative samples by retrieving top-k visually similar faces from the Laion-face-5b dataset \cite{laionface5b} utilizing CLIP \cite{clip}. Similarly, these facial images are animated by LivePortrait \cite{liveportrait} to serve as negative data, which enables the model to learn to robustly identify the personalized subjects.
In addition, these negative samples also contain relatively simple movements and other content, which inspires us to supplement them with 30 randomly selected videos from the CelebV-HQ dataset \cite{celebv} that possesses rich and high-resolution content. 
Combining them together, the personalized video chat model is equipped with robustness to both subjects and content.
% These samples serve as difficult discriminative examples for our model.

% To move the image, we animate both the positive samples (HQ-face and PhotoMaker-generated images) and negative samples using the LivePortrait model\cite{liveportrait}.
%
% The animations include five distinct facial expressions and movements: nodding, head-shaking, questioning, anger, and smiling. These videos exhibit strong temporal consistency but contain relatively simple contents, making them complementary to the data generated by ConsisID \cite{consisid}. 
%
% We also supplement our negative examples with 30 randomly selected videos from the CelebV-HQ dataset \cite{celebv}.

\begin{figure}[htb]  % !ht for "here" or "top" with high priority
    \centering
    \includegraphics[width=\columnwidth, height=1.0\linewidth]{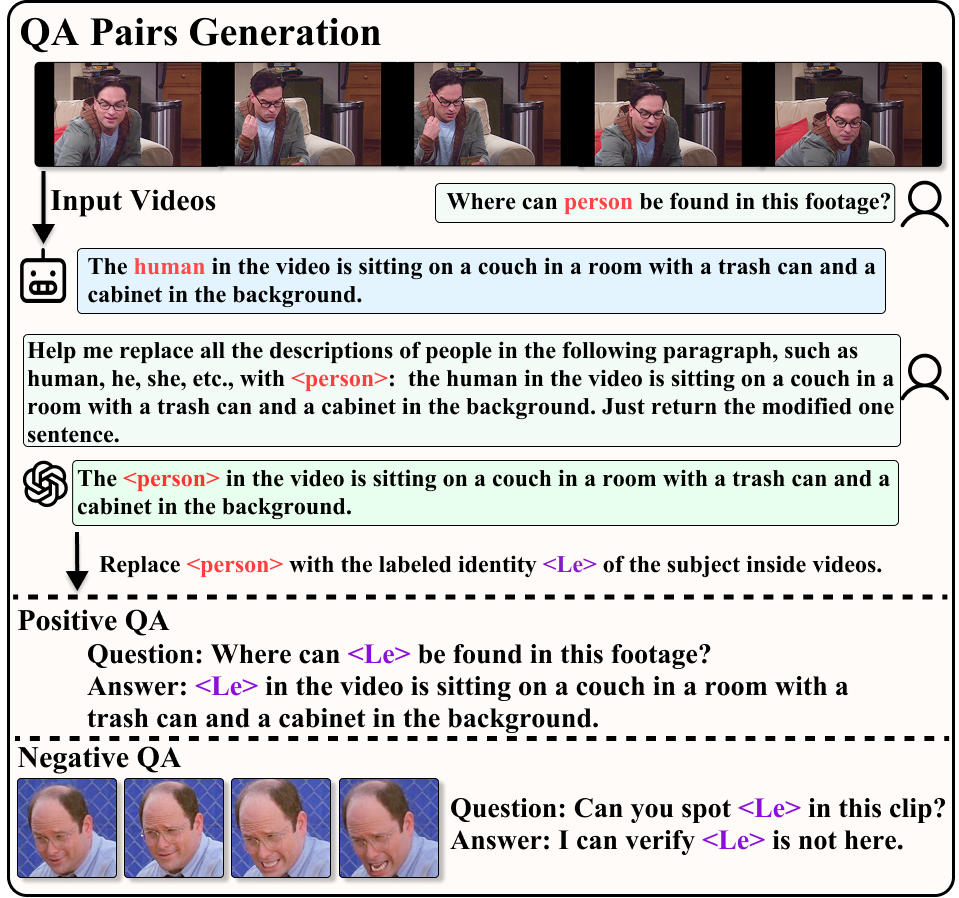}
    \caption{We illustrate the process of automatically generating question-answer pairs using InternVideo2 \cite{internvideo2} and ChatGPT \cite{achiam2023gpt}. A positive and a negative sample are shown at the bottom.}
    \label{fig:QAgen}
    %\vspace{-0.4cm}
\end{figure}

\bfsection{Question-Answer Pairs Generation}
Having various positive and negative samples, the final stage involves generating question-answer pairs using InternVideo2 \cite{internvideo2} across four categories: existence verification, appearance description, location identification, and action recognition. To enhance the linguistic naturalness, we refine these responses using ChatGPT-4o, converting generic human references into subject-specific references. The QA generation is illustrated in \cref{fig:QAgen}.

Through this comprehensive pipeline, each input natural video is transformed into $81$ different videos accompanied by $1{,}455$ question-answer pairs, significantly expanding our training dataset while maintaining subject identity consistency and guaranteeing the model's robustness.

\subsection{ReLU Routing Mixture-of-Heads Attention}
\label{ReMoH}
\bfsection{Multi-Head attention}
We begin by reviewing the standard multi-head attention mechanism \cite{mutiattention} which is based on scaled dot-product attention. For $T_1$ input tokens $\boldsymbol{X}_1 \in \mathbb{R}^{T_1 \times d_{i n}}$ where $d_{i n}$ is the input dimension, and $T_2$ tokens $\boldsymbol{X}_2 \in \mathbb{R}^{T_2 \times d_{i n}}$, attention is conducted as follows:
\begin{equation}
\begin{array}{l}
\text { Attention }(\boldsymbol{Q}, \boldsymbol{K}, \boldsymbol{V})=\operatorname{Softmax}\left(\frac{\boldsymbol{Q} \boldsymbol{K}^{\top}}{\sqrt{d_{k}}}\right) \boldsymbol{V}, \\
\text{where} \: \boldsymbol{Q}=\boldsymbol{X}_1 \boldsymbol{W}_{Q}, \boldsymbol{K}=\boldsymbol{X}_2 \boldsymbol{W}_{K}, \boldsymbol{V}=\boldsymbol{X}_2 \boldsymbol{W}_{V}.
\end{array}
\end{equation}

To enhance the representation of the attention layer, Vaswani \cite{mutiattention} proposed to use multiple attention heads to operate on the projection of the input. The transformer computes $h$ different projection of $(\boldsymbol{Q}, \boldsymbol{K}, \boldsymbol{V})$, and the concatenation form of multi-head attention is
\begin{equation}
\begin{array}{c}
\operatorname{MHA}\left(\boldsymbol{X}_1, \boldsymbol{X}_2\right)=\text { Concat }\left(\boldsymbol{H}^{1}, \boldsymbol{H}^{2}, \ldots, \boldsymbol{H}^{h}\right) \boldsymbol{W}_{O}, \\
\boldsymbol{H}^{i}=\operatorname{Attention}\left(\boldsymbol{X}_1 \boldsymbol{W}_{Q}^{i}, \boldsymbol{X}_2\boldsymbol{W}_{K}^{i}, \boldsymbol{X}_2 \boldsymbol{W}_{V}^{i}\right),
\end{array}
\end{equation}
where $\boldsymbol{W}_{Q}^{i}, \boldsymbol{W}_{K}^{i}, \boldsymbol{W}_{V}^{i}$ represent the $i_{th}$ projection matrices of the query, key and value. $\boldsymbol{W}_{O}$ is the final projection matrix,
which is decomposed by $\boldsymbol{W}_{O}=\left[\boldsymbol{W}_{O}^{1}, \boldsymbol{W}_{O}^{2}, \ldots, \boldsymbol{W}_{O}^{h}\right]$. The summation form is as follows:
\vspace{-0.2cm}
\begin{equation}
\operatorname{MHA}\left(\boldsymbol{X}_1, \boldsymbol{X}_2\right)=\sum_{i=1}^{h} \boldsymbol{H}^{i} \boldsymbol{W}_{O}^{i}.
\end{equation}

\vspace{-0.2cm}

\subsubsection{Heads as Experts with ReLU Routing}
\begin{figure}[!t]  % !ht for "here" or "top" with high priority
    \centering
    \includegraphics[width=\columnwidth,height=1.0\linewidth]{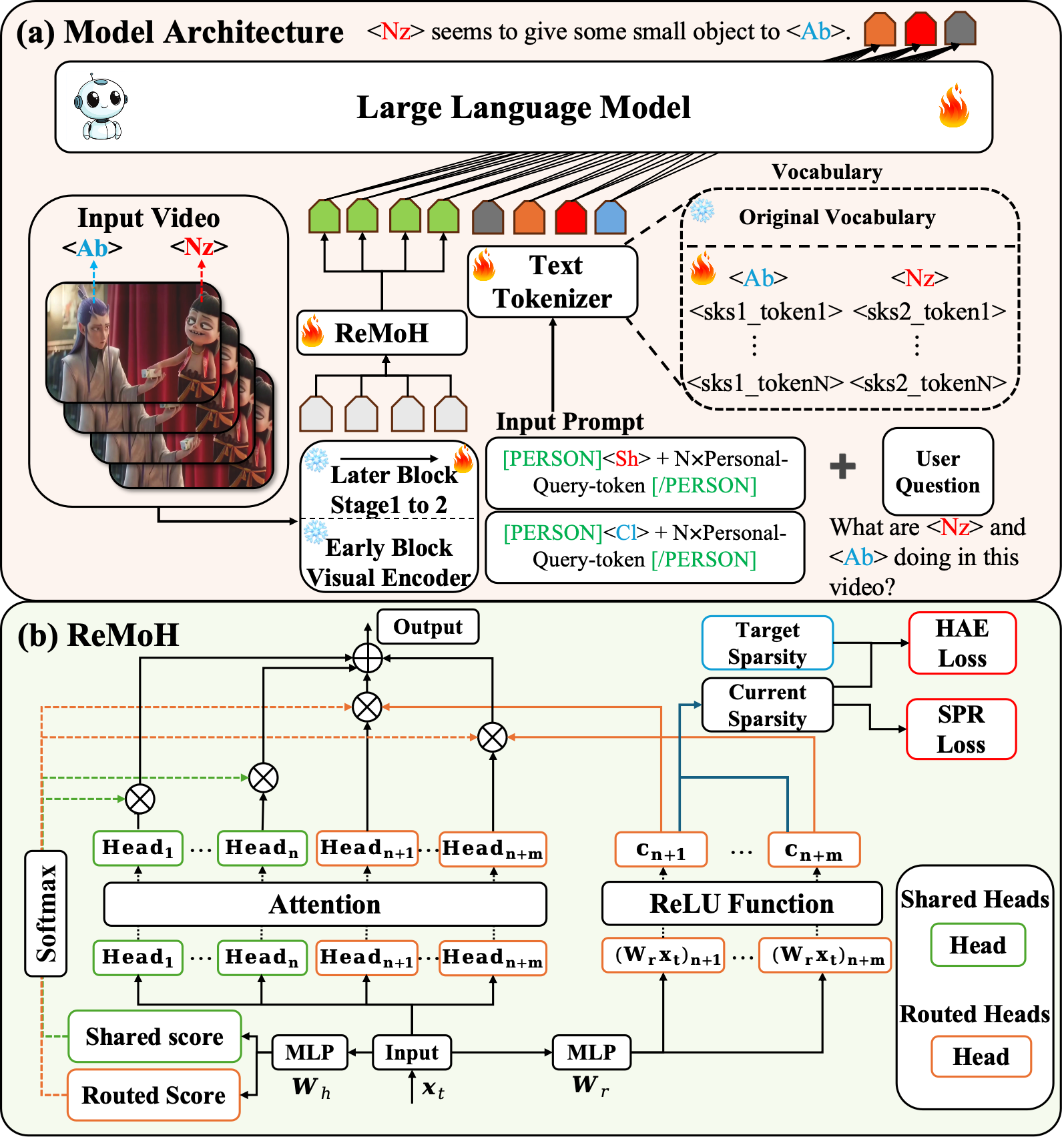}
    \caption{(a) The training pipeline of our method. (b) The proposed ReMoH technique for better specialized characteristics learning.}
    \label{fig:training}
    \vspace{-0.2cm}
\end{figure}

\bfsection{Motivation of ReMoH}
Recent study Mixture-of-Heads \cite{MOH} targets to allow each token to select the Top-k relevant heads, improving inference efficiency without sacrificing accuracy or increasing the parameters, compared with traditional multi-head attention. 
Moreover, MoH \cite{MOH} finds it can learn domain-specific
information, which is consistent with our personalized video understanding with only one-shot learning.
However, the vanilla Top-k choice of MoH limits the performance and domain-specific information learning, which is because: 
1) Top-k choice is not fully differentiable for training; 
2) The choice of heads is not adaptive and flexible, and interferes with the domain-specific knowledge learned by expert heads.
Thus, inspired by \cite{remoe}, our ReLU Routing Mixture-of-Heads Attention (ReMoH) aims at enhancing MoH with much more efficient and adaptive learning especially for the personalized subject's information, with only an increase of 2 MLP parameters. The detailed design for ReMoH can be seen in \cref{fig:training} (b).

% ReMoH's dynamic head routing mechanism allows each token to adaptively select the appropriate attention heads by a ReLU routing gate.
%
\bfsection{ReMoH with Head and ReLU Routers}
Specifically, ReMoH consists of $h$ attention heads $\boldsymbol{H}=\left\{H^{1}, H^{2}, \ldots, H^{h}\right\}$ and uses a head routers with a ReLU router to modulate the output of different expert heads. 
Formally, given input tokens $\boldsymbol{X}_1$ and $\boldsymbol{X}_2$, the output of ReMoH is the weighted sum of outputs of the heads:
\begin{equation}
\begin{array}{c}
\operatorname{ReMoH}\left(\boldsymbol{X}_1, \boldsymbol{X}_2\right)=\sum_{i=1}^{h} s_{i} \boldsymbol{H}^{i} \boldsymbol{W}_{O}^{i},
\end{array}
\end{equation}
where $s_{i}$ is the score corresponding to $H^i$. 
%
% This design enables each token to select the most relevant attention heads, boosting inference efficiency while maintaining accuracy. 
In the attention mechanism, some attention heads capture common knowledge across different contexts, such as grammatical rules in language, while others focus on context-specific knowledge  \cite{transnext}. Following MoH \cite{MOH}, we divide the heads into shared heads and routed heads, where shared heads are always activated, and routed heads will only be activated if the ReLU routing score is non-zero.
Specifically, the score for each head is defined as:
\begin{equation}
s_{i}=\left\{\begin{array}{ll}
\alpha_{1},
% \operatorname{Softmax}\left(\boldsymbol{W}_{s} \boldsymbol{x}_{t}\right)_{i},
& \text { if } 1 \leq i \leq n, \\
\alpha_{2} \operatorname{ReLU}\left(\boldsymbol{W}_{r} \boldsymbol{x}_{t}\right)_{i}, & \text { if } n < i \leq n+m,
\end{array}\right.
\end{equation}
where $n$ and $m$ are the number of shared heads and routed heads, respectively. $\boldsymbol{W}_{r} \in \mathbb{R}^{m \times d_{i n}}$ denotes the projection matrix of the ReLU router. $\alpha_{1}$ and $\alpha_{2}$, which come from the shared router, will be used to balance the contributions of the shared and routed heads, and are defined as:
\begin{equation}
\left[\alpha_{1}, \alpha_{2}\right]=\operatorname{Softmax}\left(\boldsymbol{W}_{h} \boldsymbol{x}_{t}\right).
\end{equation}

With the utility of ReLU router to choose specific routed heads to be activated, the decision process becomes fully differentiable and offers greater flexibility. 
This boosts the learning of the personalized video information, avoiding a fixed number of activated experts to limit the performance or cause computational redundancy.
A more detailed analysis can be found in our ablation study.
% The most popular method to select the most relevant expert in MOE is using top-K+softmax. but inspired by 
% Peng et al. state that the topK router will be trained in a discontinuous, non-differentiable way, limiting its performance and scalability. So we will use the ReLU to replace it, which enables more efficient dynamic allocation of computation across tokens and layers.

\subsubsection{Controlling Sparsity Strategy}
One important issue found in \cite{remoe} is that most experts are always activated during training.
Our ReMoH aims to control computation cost by managing the sparsity of the ReLU output, targeting a sparsity of $T_s=(1-\frac{b}{m})$, where $b$ is the number of routed heads we set for encouraging the activation.
To encourage the sparsity of activation, which helps to reduce computational cost and boost specific-domain learning,
we propose the Smooth Proximity Regularization (SPR) Loss to make the training more flexible:
\begin{equation}
 \mathcal{L}_{SPR}=\beta_{p} \cdot  \mathcal{L}_{Reg},
\end{equation}
where $\beta_{p}$ is a step-$p$ adaptive weight. Following prior work  \cite{lazy,prosparse}, to effectively encourage sparsity, $L_{Reg} $ always uses the $ L_1$ regularization:
\begin{equation}
 \mathcal{L}_{Reg}=\left \| \frac{1}{n}(\boldsymbol{W_{r}}\boldsymbol{x_t})  \right \|,
\end{equation}
\vspace{-0.2cm}
\begin{equation}
\beta_{p+1}=\beta_{p}\cdot e^{k \cdot(T_s-R_s) }, 
\end{equation}
where $R_s$ is the current sparsity and $R_{s}=1-\frac{1}{n}(\boldsymbol{W_{r}}\boldsymbol{x_t})$, $k$ is a scaling factor to control the change of the step-wise weight.
% can be calculated by the following format
% \begin{equation}
% R_{s}=1-\frac{1}{n}(\boldsymbol{W_{r}}\boldsymbol{x_t})
% \end{equation}
%
Such a design will make the training process more flexible, more stable, and with less extreme loss. 
%
% In the meantime, a balanced activation ratio can be guaranteed.

However, $L_{Reg} $ just focuses on increasing the sparsity, forcing the model to be more sparsely activated, which usually makes all output prone to $0$. 
Considering this, we design a Head Activation Enhancement (HAE) Loss to activate more heads to avoid some experts always being asleep:
\begin{equation}
 \mathcal{L}_{HAE}=e^{2\cdot (R_s - T_s)} - 1 \quad \text{if } R_s>T_s.
\end{equation}

$\mathcal{L}_{HAE}$ helps some heads to be more active when they fall into a state of zero. Combining the SPR and HAE, a balanced activation ratio can be guaranteed. Our final training objective is designed with the language-model used cross-entropy loss $\mathcal{L}_{LM}$:
\begin{equation}
 \mathcal{L}= \mathcal{L}_{LM}+ \mathcal{L}_{SPR}+ \mathcal{L}_{HAE}.
\end{equation}

% that activates the heads with positive output by Relu activation fuction.

% we could use the Relu to replace the conventional TopK+Softmax routing.ReMoE's continuous nature enables efficient dynamic allocation of computation across tokens and layers.
% ReLU routing function as follows:
% R(xl t) = ReLU(xl tWl)
% Given the input xl

%

%主要整合了下面两篇论文的模型并做了一些调整
%MOH: MULTI-HEAD ATTENTION AS MIXTURE-OFHEAD ATTENTION
%REMOE: FULLY DIFFERENTIABLE MIXTURE-OFEXPERTS WITH RELU ROUTING
%为了增强个性化表达学习的能力，我们设计了ReMoH(RElu Routing Multi-head Of Attention)模块，
%对于视频特征抽取中的transformer模块中的多头注意力机制模块，会分成一部分是shared的头用来保存一些共性的特征，一些是routed头用来选择一些个性化对象特征的头
%会有两个mlp，一个来得到对应的shared和routed的各自的分数
%另一个mlp来对routed的几个头的各自的分数，这里会通过relu，对于分数为负数的，相当于直接不选择了
%在原本的loss上加了两个loss
%第一个loss是当前的激活率的值*权重，权重=上一次的权重e^(k*（目标的稀疏度-当前的稀疏度）)也就是说会根据差值来调节权重的大小，但是这个由于loss是关于激活率的，很多时候经常会压制routed头的激活率很小，很多都弄成0了，并且此时的权重很大，
%一个是activation enhancement，当目前的激活率比目标的激活率低的时候，会增加一个额外的loss
%torch.exp(2 * ((目标激活率 - 当前激活率)) - 1来使得一些头的激活率重新开始，使得模型能够增加一些激活率，使得模型不会直接陷入局部最优的感觉

\subsection{Training Pipeline}
\label{training pipeline}
We implement a two-stage training strategy, transitioning from image-summary pretraining to video-QA fine-tuning, teaching the model to progressively learn both static and dynamic subject-specific features. 
This strategy adapts our \name{} model from static image understanding to dynamic video reasoning. 
The architecture is shown in \cref{fig:training}(a).
%
% This approach enables efficient learning of both subject-specific static attributes and temporal dynamics.

\bfsection{Stage 1: Image Understanding}
In the first stage, the model focuses on learning static subject-specific knowledge from separate images, where we extract the first frame of each video as input. To preserve the pre-trained visual encoder's capability, we freeze the visual encoder and only train the ReMoH components, by employing Low-Rank Adaptation (LoRA) \cite{hu2022lora} to efficiently fine-tune the Mistral-7B-Instruct-v0.3 language model \cite{3mistral7b}, which significantly reduces trainable parameters while maintaining performance.

The QA pairs used in this stage primarily consist of existence verification (e.g., ``Is \textless sks\textgreater visible in this video?") and attribute description questions (e.g.,``What is \textless sks\textgreater wearing in this video?"). These questions guide the model to focus on static subject-specific features, which helps the model to do personalized image chat.

\bfsection{Stage 2: Video Reasoning}
In the second stage, we extend the model to capture temporal dynamics and environmental interactions. We selectively train several later blocks in the visual encoder to enhance cross-frame feature integration in the personalized scenarios. This modification enables the model to recognize how subjects interact with their environments and what the subject is doing over time.

Here, we expand our QA dataset with action recognition questions (e.g., ``What movements or actions does \textless sks\textgreater  perform here?") and location identification questions (e.g., ``Can you describe \textless sks\textgreater's location relative to others?"). Utilizing both positive and negative samples for training, our \name {} model becomes robust and precise.

%这里我们分成了两步
%第一步为image-summary pretraining阶段，会从我们所有的训练视频中抽取第一帧图片来当做我们的训练输入，并且在这个阶段我们会冻结住整个visual encoder，仅仅训练我们ReMoH模块，来保证模型的学习能力。并且对于QA pairs更关注于是否存在类问题以及穿着，更关注于对象本身的特征
%第二阶段是dynamic 训练阶段，会使用所有的training的video，并且对于后半段的visual encoder（包含很多temporal的，帧与帧之间交互的层）进行训练，来增强其对个性化情况下的temporal的捕捉的能力，QA pairs中会在第一阶段的基础上添加对于所处位置以及对象的动作的QA，对于环境的交互以及人物自身的动作来进行学习，更关注一些动态的特征

%% file: 4_experiment.tex
\section{Experiments}
\label{sec:experiment}

\begin{figure*}[!t]  % !t 表示尽量将图片放在页面顶部
    \centering
    \includegraphics[width=.88\textwidth,height=.45\textwidth]{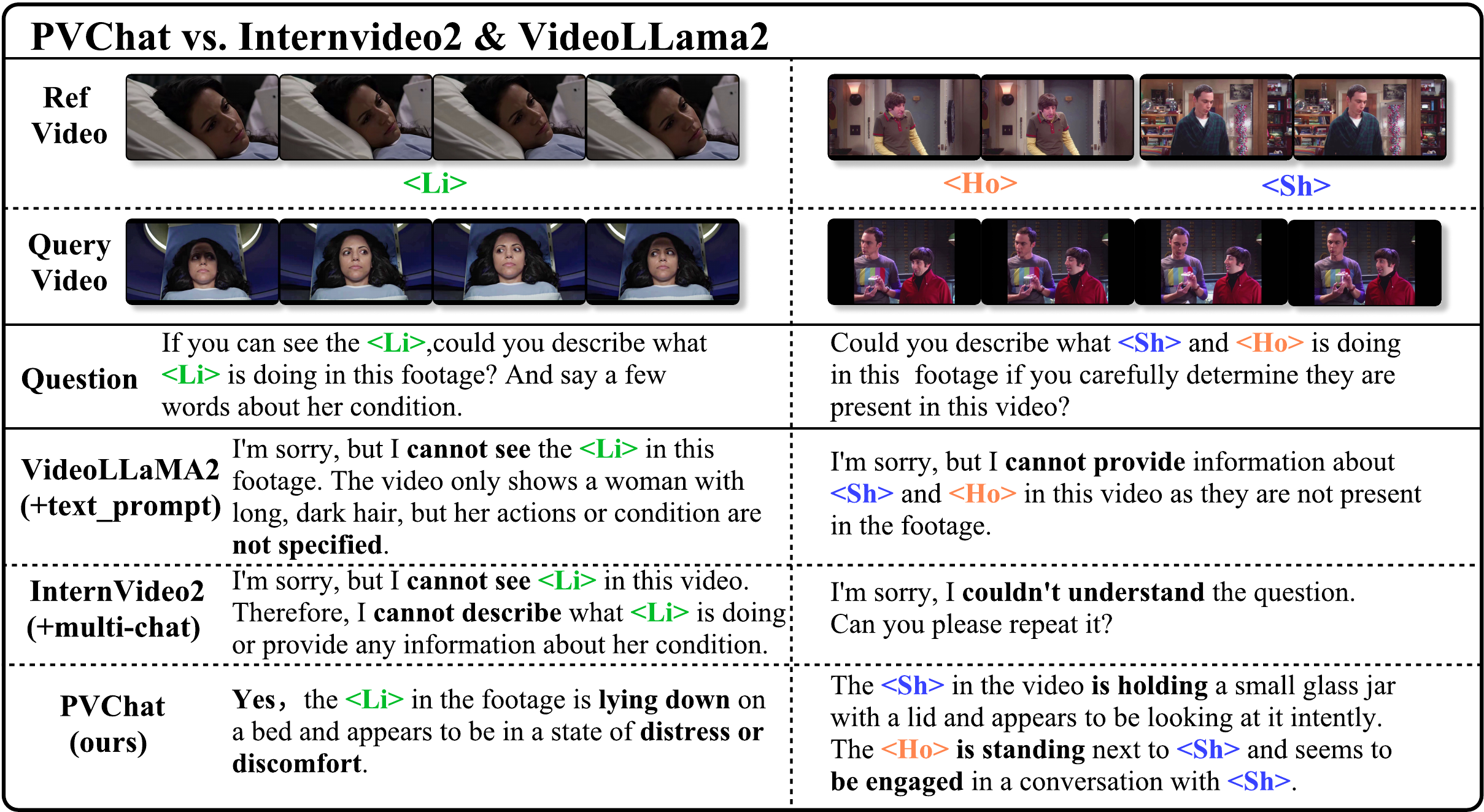}
    \caption{Examples of PVChat's ability with a learned video (e.g., a man named \textless Sh\textgreater and another man named \textless Ho\textgreater). PVChat can recognize and answer questions about the personalized concept in various scenarios, such as medical scenarios (left) and TV series (right).}
    \label{sota-2people}
    \vspace{-0.2cm}
\end{figure*}

\subsection{Settings}
\bfsection{Implementation Details} Our LLM backbone is based on the Mistral-7B-Instruct-v0.3 \cite{3mistral7b}. All models are trained on one NVIDIA L20 GPU. The first stage of training takes one epoch, and the second stage takes seven epochs. We set the batch size to $2$, the learning rates of token embedding, ReMoH, LLM Lora to $1 \times 10^{-4}$, $1 \times 10^{-5}$, and $1 \times 10^{-5}$, training time needs $3h$, video resolution is $1080 \times 1920$.For visual encoder each video will evenly sample 8 frames.

\begin{figure}[!t]  % !ht for "here" or "top" with high priority
    \centering
    \includegraphics[width=.9\columnwidth]{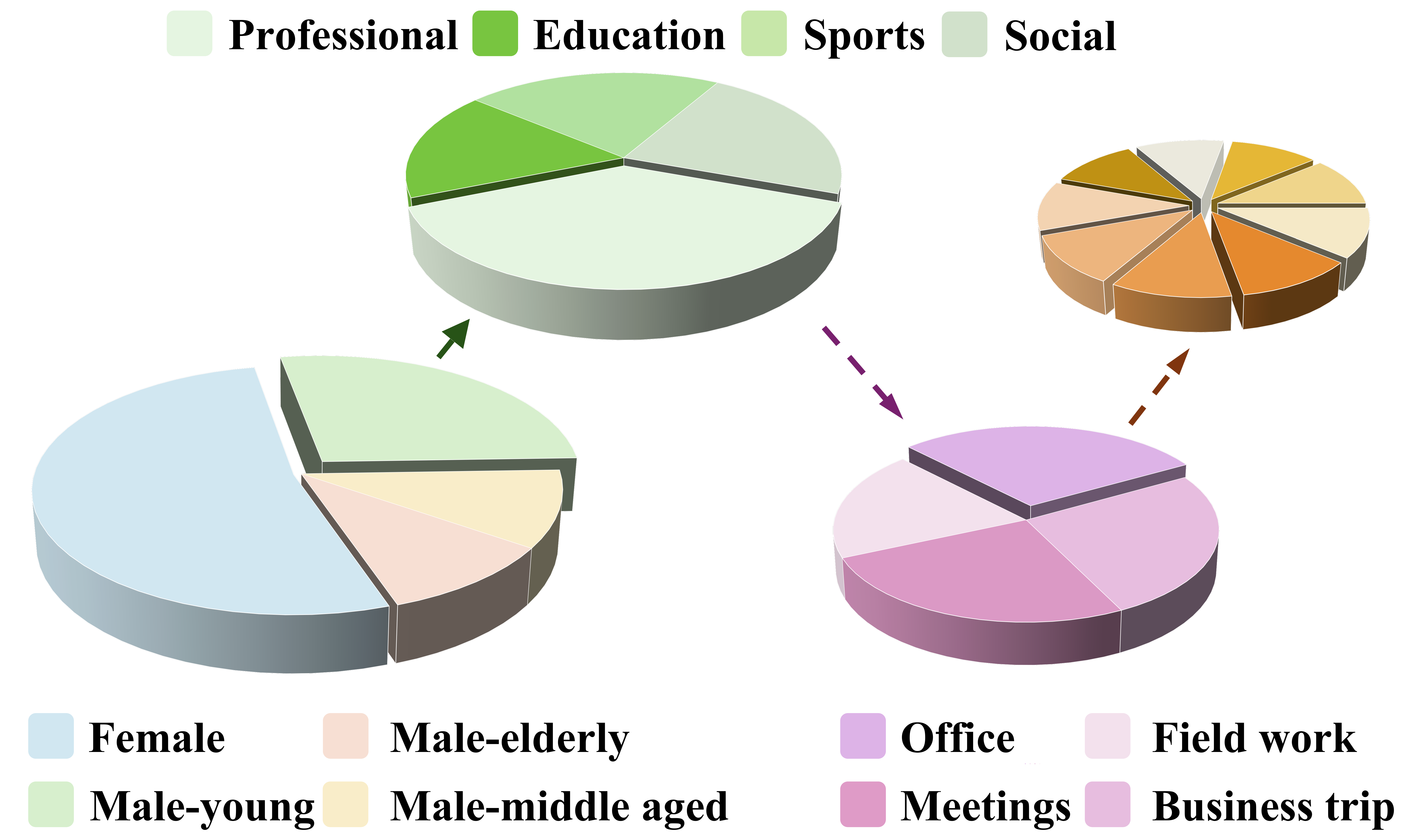}
    \caption{The hierarchical structure of our prompt library, which is carefully divided into four levels, such as gender, age, and scenarios, and provides different descriptions according to the specific subject.}
    \label{fig:bing}
    \vspace{-0.5cm}
\end{figure}

\bfsection{Datasets} To establish the dataset, we collect six different scenarios, including (Friends(6), Good Doctor(5), Ne Zha(2), doctor(3), patient(3), Big Bang(6)) $25$ characters (including doctor, patient, cartoon character, engineer, professor, etc.), and collect $304$ original videos and $2{,}304$ extended videos with over $30{,}000$ QA pairs. $300$ prompt descriptions for different scenarios are designed for different ages and genders. The structure and distribution of our dataset can be found in \cref{fig:bing}. For each character, we randomly select one video for the training set as our refer video and the other for the evaluation set as query video. We show more details and examples in our supplementary. 

\bfsection{Evaluation Metrics across Multiple Datasets} 
Following the evaluation of LLaVA \cite{llava} and Yo'LLaVA \cite{yollava}, we rigorously assess \name{} with state-of-the-art ViLLMs using five complementary metrics across four question categories: existence, appearance, action, and location queries. The final results presented represent the averaged performance across all datasets, offering a holistic view of model capabilities.

Our evaluation metrics incorporate:
(1) Accuracy (Acc) to measure the correctness of the presence or absence of subjects;
(2) BLEU and BERTScore (BS) to quantify the textual similarity between model's response and the ground truth;
(3) Entity Specificity (ES) to assess if responses contain personalized information;
(4) Descriptive Completeness (DC) to measure the quality of details inside responses.
More details about the evaluation metrics can be found in the supplementary.
%
% (1) Accuracy: Specifically designed to evaluate binary existence questions, measuring the model's ability to correctly identify the presence or absence of objects.
% (2) BLEU and BERTScore: These metrics quantify the textual similarity between generated responses and ground truth answers, capturing linguistic precision and semantic alignment.
% (3) Entity Specificity (ES): Evaluated on a 1-5 scale, this metric assesses whether responses contain personalized, contextually relevant details rather than generic statements.
% (4) Descriptive Completeness (DC): Also rated on a 1-5 scale, DC measures the logical coherence, factual correctness, and comprehensive nature of responses. This metric evaluates whether answers are thoroughly developed with appropriate supporting details and proper reasoning.
%
For appearance, action, and location queries, we employ BLEU, BS, ES, and DC to comprehensively evaluate both semantic accuracy and conversational quality of generated responses. 

\subsection{Comparison with State-of-the-Art Models}
\bfsection{Quantitative Results} To verify PVChat's effectiveness, we present the quantitative results in \cref{sota_compare}, which is compared with state-of-the-art ViLLMs, including InternVideo2 \cite{internvideo2}, VideoLLaMA2 \cite{videollama}.%, and Qwen2-7B-VL \cite{wang2024qwen2}. 
The result demonstrates that our PVChat exhibits a strong ability for personalized video understanding, offering high-accuracy and comprehensive responses. Especially for the negative sample, another model always replies that the object character is existent and no real understanding of the characteristic of the object.
Since these models do not support multi-video input in a single-round conversation, we are not able to input the reference video and query video simultaneously.
To ensure fairness, we find that InternVideo2 \cite{internvideo2} supports the multi-round conversation, allowing us to use the reference video in the first round conversation, and input our query video in the second round.
VideoLLaMA2 \cite{videollama} does not support the multi-round conversation, so we input a detailed description of the reference video, and then test the query video using the character's detailed description.

% \begin{table}

% \centering
% \caption{Quantitative evaluation of our method against four state-of-the-art methods.  Compared with 2 Sota model, our model exhibits superior performance in five metrics.}
% \label{sota_compare}
% \begin{tabular}{clllll} 
% \hline
% Mode Type    & \multicolumn{1}{c}{Acc$\uparrow$} & \multicolumn{1}{c}{BLEU$\uparrow$} & \multicolumn{1}{c}{BERTScore$\uparrow$} & \multicolumn{1}{c}{ES$\uparrow$} & \multicolumn{1}{c}{DC$\uparrow$}  \\ 
% \hline
% Internvideo2\cite{internvideo2} &                         &                          &                               &                        &                         \\
% VideoLLama2\cite{videollama}  &                         &                          &                               &                        &                         \\
% PVChat       &                         &                          &                               &                        &                         \\
% \hline
% \end{tabular}
% \end{table}
\begin{table}[t]
\centering
\vspace{-0.1cm}

\renewcommand{\arraystretch}{1.1} % 增加行间距，提高可读性
\setlength{\tabcolsep}{3pt} % 缩小列间距，使表格适应单栏
\resizebox{.9\columnwidth}{!}{ % 让表格适应单栏
\begin{tabular}{lccccc} 
\toprule
\textbf{Model Type} & \textbf{Acc$\uparrow$} & \textbf{BLEU$\uparrow$} & \textbf{BS$\uparrow$} & \textbf{ES$\uparrow$} & \textbf{DC$\uparrow$}  \\ 
\midrule
InternVideo2~\cite{internvideo2} & 0.342 & 0.046 & 0.875 & 3.041 & 1.812 \\
VideoLLaMA2~\cite{videollama}  & 0.470 & 0.082 & 0.890 & 3.012 & 3.301 \\
%Qwen-VL-7B~\cite{wang2024qwen2}  & 0.595 & 0.102 & 0.888 & 2.83 & 3.49 \\
PVChat (\textbf{Ours})      & \textbf{0.901}  & \textbf{0.562} & \textbf{0.952} & \textbf{4.940} & \textbf{4.201} \\
\bottomrule
\end{tabular}
}
\caption{Quantitative evaluation of our method against state-of-the-art methods \cite{internvideo2, videollama}. Compared with these SOTA models, our model exhibits superior performance across five metrics.}
\label{sota_compare}
\vspace{-0.3cm}
\end{table}

\bfsection{Qualitative Analysis} We conduct qualitative assessments on some challenging scenarios, including single-subject and multi-subject settings in various interactive environments and actions in \cref{sota-2people}.
% and in Fig three people(Need to fill). 
%
The result shows that our \name{} outperforms other models \cite{internvideo2, videollama} in different settings, demonstrating our superior personalized video understanding ability. Due to the shortage of personalized training and subject-specific data, previous models cannot concisely capture personalized information and show reasonable responses with correct identity.

\begin{figure*}[!t]
    \centering
    \includegraphics[width=.84\textwidth, height=.27\textwidth]{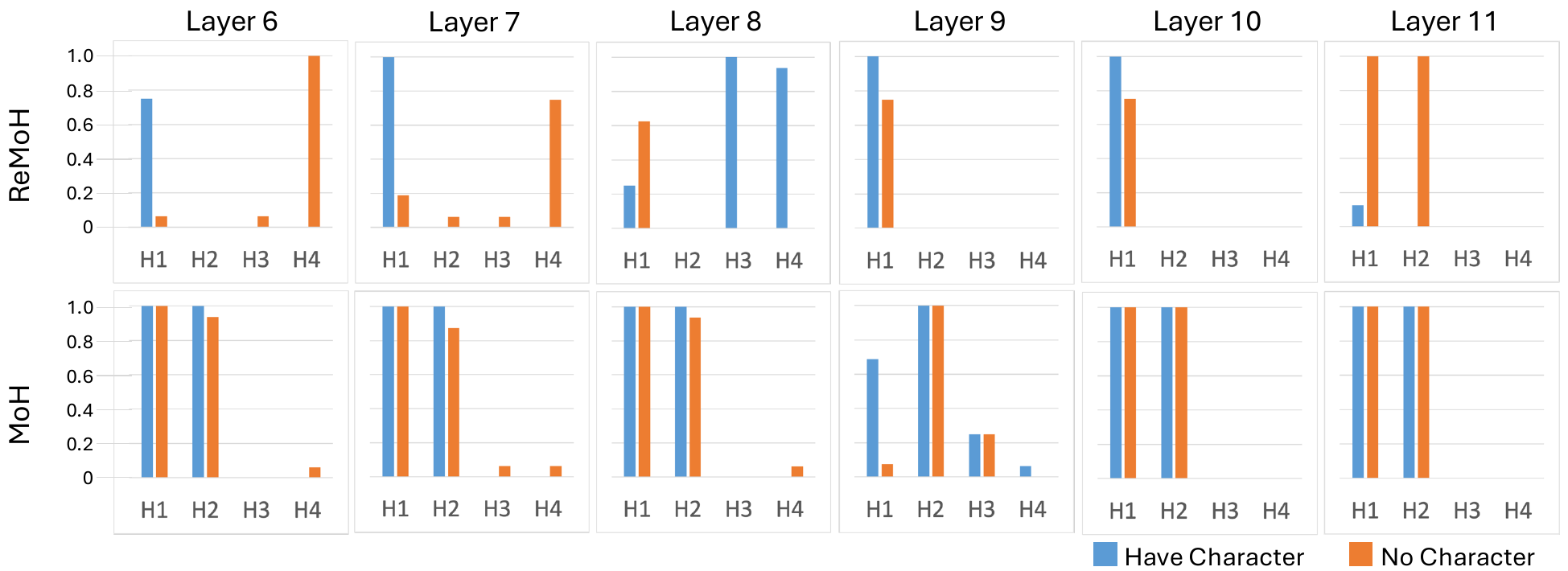}
    \vspace{-0.2cm}
    \caption{The comparison of expert heads activation between MoH \cite{MOH} and ReMoH in different layers, where $\text{H}_{i}$ represents the $i^{th}$ head. Orange refers to the video without the target individual, while blue represents the video having the character.}
    \label{compare moh and ReMoH}
    \vspace{-0.28cm}
\end{figure*}

\subsection{Ablation Study}

\bfsection{Analysis of ReMoH} To explore the impact of ReMoH, we conduct experiments as shown in \cref{ablation_each_content} and visualize the heads of expert activation ratio in \cref{compare moh and ReMoH}. We choose the typical transformer architecture Q-former \cite{Qformer} with the LLM backbone \cite{3mistral7b} as our baseline.
Compared with the baseline and incorporating the typical MoH \cite{MOH}, our ReMoH results in respective improvements of $18.6\%$, $9.8\%$ on the Acc, where our ReMoH significantly enhances the model's capability of personalizing understanding and chatting.

In \cref{compare moh and ReMoH}, we find that
% ReLU changes the softmax+topk will effectively promote 
ReMoH effectively allocates some certain heads of expert to extract the video feature about the target character, while MoH \cite{MOH} almost evenly distributes expert heads to learn the knowledge. 
For example, if using MoH \cite{MOH}, in most layers, no matter the character appears in the video or not, the first and second heads are always activated for video reasoning, verifying that these heads of expert actually do not learn about specific information.
On the contrary, with the utility of ReMoH, in most layers it is observed that the activation of expert heads undergoes significant changes from no characters to having characters.
This pattern supports that the model can learn personalized knowledge with ReMoH better, improving the QA performance related to specific identity, as presented in \cref{ablation_each_content}. 

% For the test video, whether the target person appears different, ReMoH can make corresponding different choices to select the more appropriate head activation, thus promoting better personalized feature extraction than MOH(Just choose similar head to activate).

% \begin{table}[H]
% \centering
% \caption{Ablation study about each content in \textless NZ\textgreater}
% \label{ablation_each_content}
% \begin{tabularx}{\columnwidth}{lXXXXX} 
% \hline
%                & Acc$\uparrow$     & BLUE$\uparrow$    & BS$\uparrow$  & ES$\uparrow$    & DC$\uparrow$     \\ 
% \hline
% baseline       &        &        &           &      &       \\
% baseline+MOH   & 0.881 & 0.485 & 0.942    & 5.00    & 3.37  \\
% baseline+ReMoH & 0.916 & 0.489 & 0.941    & 4.95 & 4.18  \\
% \hline
% \end{tabularx}
% \end{table}
\begin{table}[tb]
\centering
% \vspace{-0.3 cm}
\renewcommand{\arraystretch}{1.} % 增加行间距
\setlength{\tabcolsep}{3pt} % 减少列间距，使表格更紧凑
\resizebox{0.85\columnwidth}{!}{ % 让表格适应单栏
\begin{tabular}{lccccc} 
\toprule
\textbf{Method} & \textbf{Acc$\uparrow$} & \textbf{BLEU$\uparrow$} & \textbf{BS$\uparrow$} & \textbf{ES$\uparrow$} & \textbf{DC$\uparrow$}  \\ 
\midrule
Baseline       & 0.733    & 0.550    &0.904        & 4.735    & 4.142   \\
Baseline + MoH \cite{MOH}   & 0.813 & 0.558 & 0.926    & 4.939 & 4.191  \\
Baseline + ReMoH & \textbf{0.901} & \textbf{0.562} & \textbf{0.952}    & \textbf{4.940}  & \textbf{4.201}  \\
\bottomrule
\end{tabular}
}
\vspace{-0.2cm}
\caption{Ablations on the proposed ReMoH, where ReMoH significantly outperforms MoH in Acc.}
% \vspace{-0.1cm}
\label{ablation_each_content}
\vspace{-0.5cm}
\end{table}

\bfsection{Influence of SPR and HAE}
We conducted ablations to verify the effectiveness of SPR and HAE losses. As shown in \cref{ablation_two_loss}, the introduction of SPR successfully smooths loss fluctuations and enhances training stability compared with MoH \cite{MOH}. However, models with only SPR suffer from very low expert head activation ratio, limiting personalized feature extraction. Adding HAE loss alongside SPR, the head activation diversity increases significantly, allowing the model to capture more nuanced personalized representations. 

% This combination created a synergistic effect that substantially improved personalized dialogue capabilities across all evaluation metrics.

\begin{table}[tb]
\centering
% \vspace{-0.2cm}
\renewcommand{\arraystretch}{1.} % 增加行间距
\setlength{\tabcolsep}{3pt} % 减少列间距，使表格更紧凑
\resizebox{0.9\columnwidth}{!}{ % 让表格适应单栏
\begin{tabular}{lccccccc} 
\toprule
\textbf{Method} & \textbf{Activate Rate} & \textbf{loss$\downarrow$}  & \textbf{Acc$\uparrow$} & \textbf{BLEU$\uparrow$} & \textbf{BS$\uparrow$} & \textbf{ES$\uparrow$} & \textbf{DC$\uparrow$}  \\ 
\midrule
PVChat \textit{w/o} SPR and MAE       & --    & nan    &--        &--   & --  &--   & --  \\
PVChat \textit{w/o} MAE   & 0.217 & 0.085 & 0.746    & 0.555  & 0.926 & 4.913  & 4.112 \\
PVChat & \textbf{0.552} & \textbf{0.028} & \textbf{0.901}    & \textbf{0.562}  & \textbf{0.952}& \textbf{4.940}  & \textbf{4.201}  \\
\bottomrule
\end{tabular}
}
\vspace{-0.1cm}
\caption{Ablations on SPR and HAE losses. It verifies that SPR and HAE guarantee stability and enhance learning of the expert heads.}
% \vspace{-0.1cm}
\label{ablation_two_loss}
\vspace{-0.2cm}
\end{table}

\bfsection{Contribution of Each Type of Data} We conduct an ablation on the dataset creation. \cref{ablation_datasets}
presents all metrics for the personalization conversion. Using vanilla datasets (with only one positive sample) fails to perform well on personalized questions, always responding with ``yes'' to all questions. After injecting the negative samples, it could recognize the identity of the character to a certain degree, but still suffers from some complex questions about action and location. After training with the generated data from ConsisID \cite{consisid} and LivePortrait \cite{liveportrait}, the model is capable of understanding the specific information more accurately and robustly.
% \begin{table}[H]
% \centering
% \caption{Ablation study about the dataset creation in \textless Le\textgreater}
% \label{ablation_datasets}
% \begin{tabularx}{\columnwidth}{llllll} 
% \hline
% \multicolumn{1}{c}{Data Type}    & \multicolumn{1}{c}{Acc$\uparrow$} & \multicolumn{1}{c}{BLEU$\uparrow$ } & \multicolumn{1}{c}{BS$\uparrow$ } & \multicolumn{1}{c}{ES$\uparrow$ } & \multicolumn{1}{c}{DC$\uparrow$ }  \\ 
% \hline
% one positive                     &                         &                          &                               &                        &                         \\
% +Negitive                        &      0.807                    &   0.428                        &                       0.940        &              4.92          &                  4.25       \\
% +Consisid Positive     &   0.9037                    &          0.593               &                    0.9527        &            4.908      &   4.416                 \\ 

% +LivePortrait Positive &0.948                         &  0.597                        &            0.955                   &           4.986             & 4.420        \\
% \hline
% \end{tabularx}
% \end{table}
\begin{table}[tb]
\centering
% \vspace{-0.2cm}
\renewcommand{\arraystretch}{1.1} % 增加行距，提升可读性
\setlength{\tabcolsep}{3pt} % 减少列间距，使表格更紧凑
\resizebox{0.9\columnwidth}{!}{ % 让表格自动适应单栏
\begin{tabular}{lccccc} 
\toprule
\textbf{Data Type} & \textbf{Acc$\uparrow$} & \textbf{BLEU$\uparrow$} & \textbf{BS$\uparrow$} & \textbf{ES$\uparrow$} & \textbf{DC$\uparrow$}  \\ 
\midrule
one positive                     & 0.464 & 0.417 & 0.905 & 4.826 & 3.947 \\
+Negative                        & 0.584  & 0.418  & 0.931  & 4.899  & 4.120  \\
+ConsisID Positive     & 0.781  & 0.532  & 0.927  & 4.929  & 4.132  \\ 
+LivePortrait Positive & \textbf{0.901}  & \textbf{0.562}  & \textbf{0.952}  & \textbf{4.940}  & \textbf{4.201}  \\
\bottomrule
\end{tabular}
}
\vspace{-0.1cm}
\caption{Ablations on the dataset collection, where combining all types of designed data, the model performs accurately and robustly.}
% \vspace{-0.1cm}
\label{ablation_datasets}
\vspace{-0.4cm}
\end{table}

\bfsection{Influence of Token Numbers} We set each question prompt corresponding to $N=16$ tokens per character. 
% If $N=0$, training is with only the \textless sks\textgreater token. 
As the token number increases, the metrics overall increase, but decrease when larger than 16. This is because too many tokens make the model hard to capture the characteristics of the subject. The detailed results can be found in our supplementary.

% \begin{table}[H]
% \centering
% \vspace{-0.2cm}
% \renewcommand{\arraystretch}{.9} % 控制行间距
% \setlength{\tabcolsep}{3pt} % 控制列间距
% \resizebox{0.8\columnwidth}{!}{ % 这里调整 0.85 可缩放表格整体大小
% \begin{tabular}{cccccc} 
% \toprule
% \textbf{Number} & \textbf{Acc$\uparrow$} & \textbf{BLEU$\uparrow$} & \textbf{BS$\uparrow$} & \textbf{ES$\uparrow$} & \textbf{DC$\uparrow$}  \\ 
% \midrule
% 0  & 0.801  & 0.495  & 0.939  & 5.00  & 4.00  \\
% 4  & 0.871  & 0.592  & 0.951  & 5.00  & 4.50  \\
% 8  & 0.890  & 0.553  & 0.949  & 5.00  & \textbf{4.58}  \\
% 12 & 0.895  & 0.564  & 0.949  & 5.00  & 4.41  \\
% 16 & \textbf{0.922}  & \textbf{0.606}  & \textbf{0.952}  & 4.74  & 4.38  \\
% 20 & 0.882  & 0.554  & 0.949  & 4.96  & 4.25  \\
% \bottomrule
% \end{tabular}
% }
% \vspace{-0.3cm}
% \caption{Ablations on the number of tokens.}
% % \vspace{-0.1cm}
% \label{ablation_token_number}
% \vspace{-0.4cm}
% \end{table}

%% file: 5_conclusion.tex
\section{Conclusion}
\label{sec:conclusion}
This paper introduces \name{}, the first personalized ViLLM that enables personalized subject learning from a single video. We collect comprehensive datasets over $2{,}300$ videos, propose ReMoH for specific knowledge learning, and employ a two-stage training strategy from image pre-training to video fine-tuning, with SPR and HAE losses. Evaluations show the superiority of our model in personalized video QA, making it potential for smart healthcare and home scenarios.

%% file: supp.tex
In this supplemental material, the readers can find: 

\begin{itemize}
\item Experimental details about the ablation study of token number;
\item Details about multiple metrics utilized for personalized QA quality;
\item More examples of recognition question answering;
\item More experiment detail about Multi Character training
\item Some examples of existential questions and answers for single and two entities;
\item Presentation of templates for GPT prompt queries;
\item Presentation of 25 characters;

\end{itemize}

\section{Ablation Study About the Token Number}

We set each question prompt corresponding to $N=16$ tokens per character. 
If $N=0$, training is only with the \textless sks\textgreater token. 
As the token number increases, the metrics overall increase. However, when the number is larger than $16$, the performance decreases. This is because too many tokens can make the model hard to capture the characteristics of the subject. The detailed experimental results are shown in \cref{ablation_token_number}.

\begin{table}[htb]
\centering
% \vspace{-0.2cm}
\renewcommand{\arraystretch}{.9} % 控制行间距
\setlength{\tabcolsep}{3pt} % 控制列间距
\resizebox{0.8\columnwidth}{!}{ % 这里调整 0.85 可缩放表格整体大小
\begin{tabular}{cccccc} 
\toprule
\textbf{Number} & \textbf{Acc$\uparrow$} & \textbf{BLEU$\uparrow$} & \textbf{BS$\uparrow$} & \textbf{ES$\uparrow$} & \textbf{DC$\uparrow$}  \\ 
\midrule
0  & 0.801  & 0.495  & 0.939  & 5.00  & 4.00  \\
4  & 0.871  & 0.592  & 0.951  & 5.00  & 4.50  \\
8  & 0.890  & 0.553  & 0.949  & 5.00  & \textbf{4.58}  \\
12 & 0.895  & 0.564  & 0.949  & 5.00  & 4.41  \\
16 & \textbf{0.922}  & \textbf{0.606}  & \textbf{0.952}  & 4.74  & 4.38  \\
20 & 0.882  & 0.554  & 0.949  & 4.96  & 4.25  \\
\bottomrule
\end{tabular}
}
% \vspace{-0.3cm}
\caption{Ablations on the number of tokens in \textless Aa\textgreater.}
% \vspace{-0.1cm}
\label{ablation_token_number}
% \vspace{-0.4cm}
\end{table}

\section{Details of the Evaluation Metrics}
Here we add more details of the metrics which are utilized to measure the quality of personalized video chat: 

(1) Accuracy: Specifically designed to evaluate binary existence questions, measuring the model's ability to correctly identify the presence or absence of objects. 

(2) BLEU and BERTScore: These metrics quantify the textual similarity between generated responses and ground truth answers, capturing linguistic precision and semantic alignment.

(3) Entity Specificity (ES): Evaluated on a 1-5 scale, this metric assesses whether responses contain personalized, contextually relevant details rather than generic statements.

(4) Descriptive Completeness (DC): Also rated on a 1-5 scale, DC measures the logical coherence, factual correctness, and comprehensive nature of responses. This metric evaluates whether answers are thoroughly developed with appropriate supporting details and proper reasoning.

\section{More Examples of Our PVChat Model}
\begin{figure*}[htb]  % !t 表示尽量将图片放在页面顶部
    \centering
    \includegraphics[width=.88\textwidth]{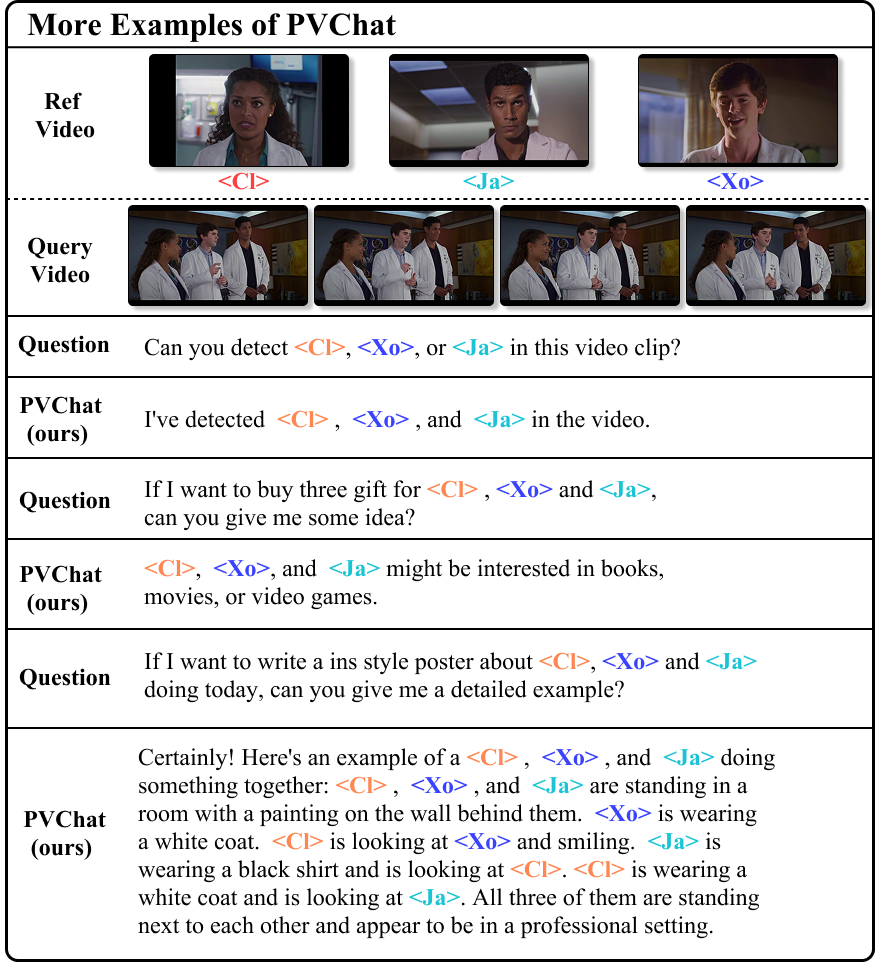}
    \caption{Example of PVChat.}
    \label{supp_figtop}
    % \vspace{-0.2cm}
\end{figure*}

\begin{figure*}[htb]  % !t 表示尽量将图片放在页面顶部
    \centering
    \includegraphics[width=.88\textwidth,height=.529\textwidth]{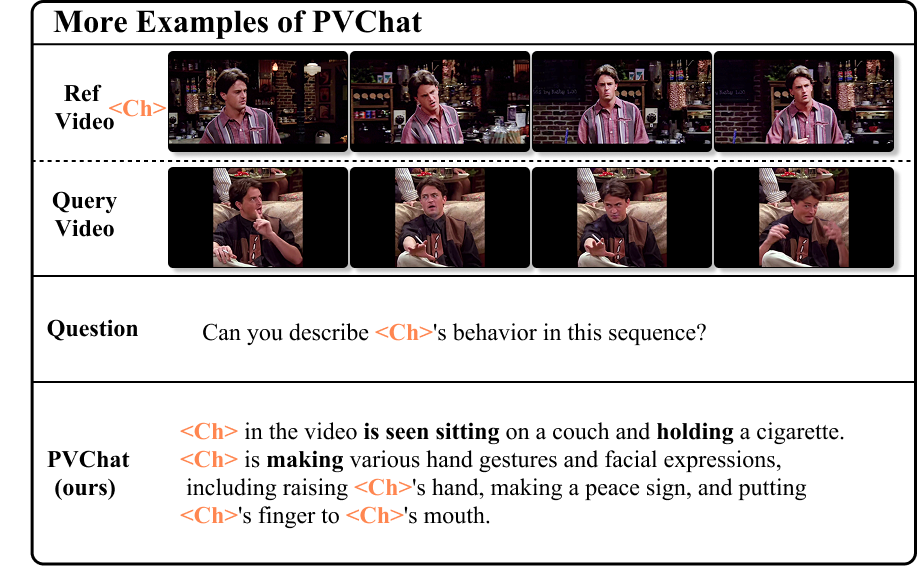}
    \caption{Example of PVChat.}
    \label{supp_figt1}
    % \vspace{-0.2cm}
\end{figure*}

\begin{figure*}[htb]  % !t 表示尽量将图片放在页面顶部
    \centering
    \includegraphics[width=.88\textwidth,height=.529\textwidth]{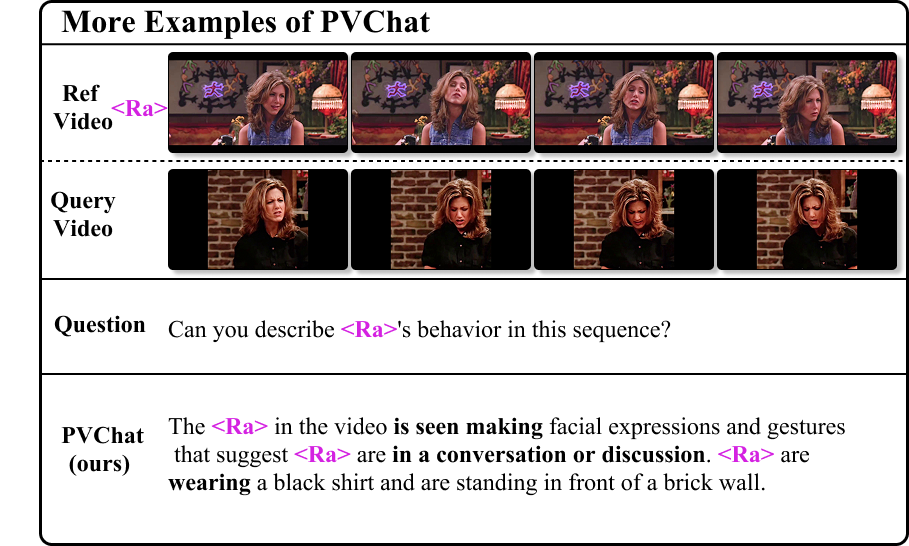}
    \caption{Example of PVChat.}
    \label{supp_fig2}
    % \vspace{-0.2cm}
\end{figure*}

\begin{figure*}[htb]  % !t 表示尽量将图片放在页面顶部
    \centering
    \includegraphics[width=.88\textwidth,height=.529\textwidth]{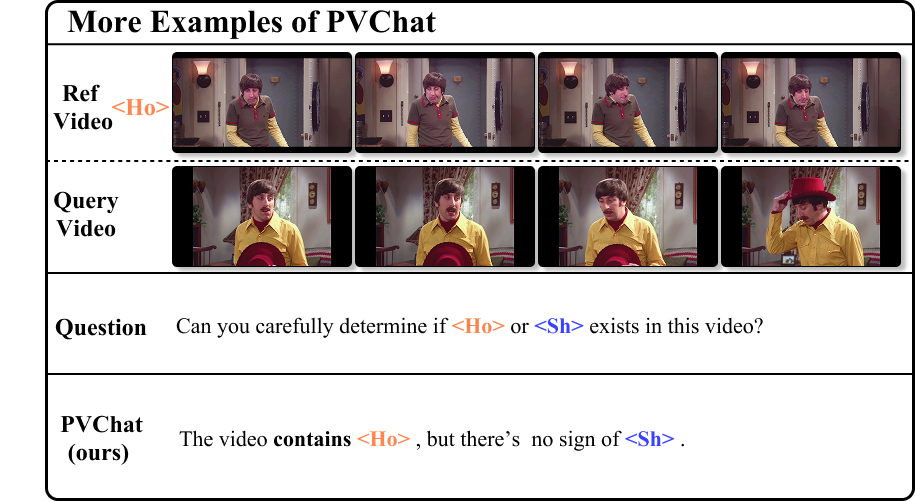}
    \caption{Example of PVChat.}
    \label{supp_fig3}
    % \vspace{-0.2cm}
\end{figure*}

\begin{figure*}[htb]  % !t 表示尽量将图片放在页面顶部
    \centering
    \includegraphics[width=.88\textwidth,height=.529\textwidth]{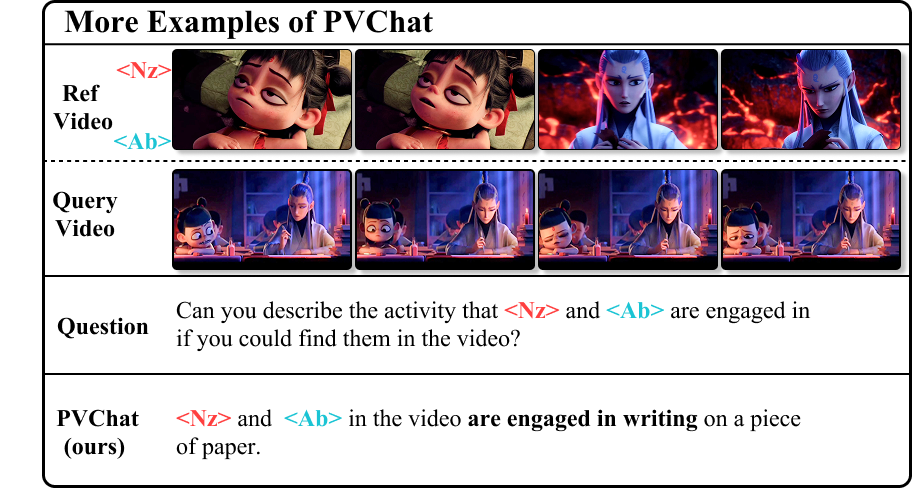}
    \caption{Example of PVChat.}
    \label{supp_fig4}
    % \vspace{-0.2cm}
\end{figure*}

From \cref{supp_figtop} to \cref{supp_fig4}, more personalized video chat examples of various scenarios are displayed. For example, in \cref{supp_figtop}, there are three characters named \textless Cl\textgreater, \textless Ja\textgreater and \textless Xo\textgreater in a laboratory scene, which is quite challenging for video understanding. Our PVChat not only accurately identifies these three men by their names but also gives reasonable suggestions for some additional questions about giving gifts. What's more, when the model is asked about their behavior, it successfully captures their locations and can do reasoning to guess this is a professional setting. This demonstrates that our model has strong personalized reasoning capability, even in a multiple subject setting. More details can be seen in the figures.

\section{More experiment detail about Multi-Character training}
\textbf{Epoch} One person training needs 8 epochs, and two people training need 16 epochs, and three people training need 24 epochs.

\noindent\textbf{ReMoH} Within the ReMoH architecture, we implement cross-attention mechanisms at alternating layers to facilitate interaction between video embeddings (extracted by the visual encoder) and query embeddings. Our approach adaptively scales with the number of subjects: for single-person training, we incorporate an additional $16 \times \text{token\_length} \times 768$ dimensional feature space; for two-person training, this expands to $32 \times \text{token\_length} \times 768$ and for three-person training, it further increases to $48 \times \text{token\_length} \times 768$.

Crucially, these query tokens are associated with corresponding masks that are only deactivated during the training of their respective subject videos. This selective masking strategy ensures that each query embedding specifically attends to its designated subject, effectively preventing cross-contamination between different identity representations.

\noindent\textbf{Special tokens} For multi-person scenarios, we employ the special tokens [PERSON] and [\textbackslash PERSON] as delimiters to clearly delineate the boundary between one person's SKS tokens and another's. This demarcation approach enables the model to accurately distinguish when the sequence of tokens associated with one individual ends and those belonging to another begin.

\section{Example of 4 Different Questions}
From \cref{tab:single_entity_yes} to \cref{tab:two_entities_mixed}, the detailed existence question for one entity and two entities are displayed. We show the other three questions about the appearance, action, and location in \cref{tab:single 3question}.
% --- Table 1: Single Entity (Yes Answers) ---

\begin{table*}[t]
    \centering
    \captionsetup{justification=centering}
    \renewcommand{\arraystretch}{1.4} % Increase row height
    \small % Reduce text size
    \begin{tabular}{p{1.5cm} p{6.5cm} p{5.5cm}}  % Manually adjusted column widths
        \toprule
        \textbf{Index} & \textbf{Question} & \textbf{Yes Answer} \\
        \midrule
        1  & Is there any trace of \textless sks\textgreater\ in this footage? & Yes, \textless sks\textgreater\ is in this video. \\
        2  & Can you detect \textless sks\textgreater\ in this video clip? & I can confirm that \textless sks\textgreater\ appears. \\
        3  & Does \textless sks\textgreater\ show up anywhere in this recording? & \textless sks\textgreater\ is present in this recording. \\
        4  & Is \textless sks\textgreater\ visible in this video? & The video contains \textless sks\textgreater. \\
        5  & Could you verify if \textless sks\textgreater\ is here? & I've identified \textless sks\textgreater. \\
        6  & Does this footage include \textless sks\textgreater? & \textless sks\textgreater\ is shown in this video. \\
        7  & Can you spot \textless sks\textgreater\ in this clip? & Yes, \textless sks\textgreater\ appears here. \\
        8  & Is \textless sks\textgreater\ present in this video? & I can verify that \textless sks\textgreater\ is present. \\
        9  & Does \textless sks\textgreater\ appear in this footage? & The footage shows \textless sks\textgreater. \\
        10 & Can you tell if \textless sks\textgreater\ is shown here? & \textless sks\textgreater\ is in this video clip. \\
        11 & Is \textless sks\textgreater\ in this video segment? & I've detected \textless sks\textgreater. \\
        12 & Can you confirm \textless sks\textgreater's presence? & Yes, \textless sks\textgreater\ is featured. \\
        13 & Does this clip contain \textless sks\textgreater? & The video includes \textless sks\textgreater. \\
        14 & Is \textless sks\textgreater\ featured in this recording? & I can see \textless sks\textgreater. \\
        15 & Can you find \textless sks\textgreater\ in this video? & \textless sks\textgreater\ is definitely here. \\
        16 & Is \textless sks\textgreater\ shown in any frame? & Yes, I've found \textless sks\textgreater. \\
        17 & Does this video show \textless sks\textgreater? & This video shows \textless sks\textgreater. \\
        18 & Is \textless sks\textgreater\ visible anywhere? & \textless sks\textgreater\ is visible. \\
        19 & Can you see \textless sks\textgreater? & Yes, \textless sks\textgreater\ has been captured. \\
        20 & Is \textless sks\textgreater\ in this video? & The video clearly shows \textless sks\textgreater. \\
        21 & Can you recognize \textless sks\textgreater? & I've spotted \textless sks\textgreater. \\
        22 & Does \textless sks\textgreater\ appear at all? & \textless sks\textgreater\ appears in this video. \\
        23 & Is \textless sks\textgreater\ recorded here? & Yes, this footage contains \textless sks\textgreater. \\
        24 & Can you identify \textless sks\textgreater? & I can recognize \textless sks\textgreater. \\
        25 & Is \textless sks\textgreater\ present? & \textless sks\textgreater\ is clearly visible. \\
        \bottomrule
    \end{tabular}
    \caption{ Example of positive existence verification answering (Single entity).}
    \label{tab:single_entity_yes}
    
\end{table*}

% --- Table 2: Single Entity (No Answers) ---

\begin{table*}[t]
    \centering
    \captionsetup{justification=centering}
    \renewcommand{\arraystretch}{1.4}
    \small
    \begin{tabular}{p{1.5cm} p{6.5cm} p{6.5cm}}  
        \toprule
        \textbf{Index} & \textbf{Question} & \textbf{No Answer} \\
        \midrule
        1  & Is there any trace of \textless sks\textgreater\ in this footage? & No, \textless sks\textgreater\ is not in this video. \\
        2  & Can you detect \textless sks\textgreater\ in this video clip? & I cannot detect \textless sks\textgreater. \\
        3  & Does \textless sks\textgreater\ show up anywhere in this recording? & This video does not contain \textless sks\textgreater. \\
        4  & Is \textless sks\textgreater\ visible in this video? & \textless sks\textgreater\ is not shown. \\
        5  & Could you verify if \textless sks\textgreater\ is here? & There is no sign of \textless sks\textgreater. \\
        6  & Does this footage include \textless sks\textgreater? & \textless sks\textgreater\ does not appear. \\
        7  & Can you spot \textless sks\textgreater\ in this clip? & I can confirm \textless sks\textgreater\ is not here. \\
        8  & Is \textless sks\textgreater\ present in this video? & The footage does not include \textless sks\textgreater. \\
        9  & Does \textless sks\textgreater\ appear in this footage? & There's no evidence of \textless sks\textgreater. \\
        10 & Can you tell if \textless sks\textgreater\ is shown here? & \textless sks\textgreater\ is not in this video. \\
        11 & Is \textless sks\textgreater\ in this video segment? & I've checked, \textless sks\textgreater\ is not present. \\
        12 & Can you confirm \textless sks\textgreater's presence? & This video does not show \textless sks\textgreater. \\
        13 & Does this clip contain \textless sks\textgreater? & I see no sign of \textless sks\textgreater. \\
        14 & Is \textless sks\textgreater\ featured in this recording? & \textless sks\textgreater\ is absent. \\
        15 & Can you find \textless sks\textgreater\ in this video? & The video does not show \textless sks\textgreater. \\
        16 & Is \textless sks\textgreater\ shown in any frame? & I cannot find \textless sks\textgreater. \\
        17 & Does this video show \textless sks\textgreater? & \textless sks\textgreater\ is not visible. \\
        18 & Is \textless sks\textgreater\ visible anywhere? & I can verify \textless sks\textgreater\ is not here. \\
        19 & Can you see \textless sks\textgreater? & The video has no \textless sks\textgreater. \\
        20 & Is \textless sks\textgreater\ in this video? & \textless sks\textgreater\ does not exist in this video. \\
        21 & Can you recognize \textless sks\textgreater? & I find no trace of \textless sks\textgreater. \\
        22 & Does \textless sks\textgreater\ appear at all? & This clip does not contain \textless sks\textgreater. \\
        23 & Is \textless sks\textgreater\ recorded here? & \textless sks\textgreater\ is not present. \\
        24 & Can you identify \textless sks\textgreater? & I cannot identify \textless sks\textgreater. \\
        25 & Is \textless sks\textgreater\ present? & There is no \textless sks\textgreater\ here. \\
        \bottomrule
    \end{tabular}
    \caption{ Example of negative existence verification answering (Single entity).}
    \label{tab:single_entity_no}
\end{table*}

% --- Table 3: Two Entities (Both Yes) ---

\begin{table*}[t]
    \centering
    \captionsetup{justification=centering}
    \renewcommand{\arraystretch}{1.4}
    \small
    \begin{tabular}{p{1.5cm} p{6.5cm} p{6.5cm}}  
        \toprule
        \textbf{Index} & \textbf{Question} & \textbf{Both Yes Answer} \\
        \midrule
        1  & Is there any trace of \textless sks1\textgreater\ or \textless sks2\textgreater\ in this footage? & Both \textless sks1\textgreater\ and \textless sks2\textgreater\ are present in this video. \\
        2  & Can you detect \textless sks1\textgreater\ or \textless sks2\textgreater\ in this video clip? & I can detect both \textless sks1\textgreater\ and \textless sks2\textgreater\ in the footage. \\
        3  & Do \textless sks1\textgreater\ or \textless sks2\textgreater\ show up anywhere in this recording? & The video shows both \textless sks1\textgreater\ and \textless sks2\textgreater\ clearly. \\
        4  & Are \textless sks1\textgreater\ or \textless sks2\textgreater\ visible in this video? & \textless sks1\textgreater\ and \textless sks2\textgreater\ are both visible in this recording. \\
        5  & Could you verify if \textless sks1\textgreater\ or \textless sks2\textgreater\ are here? & I've identified both \textless sks1\textgreater\ and \textless sks2\textgreater\ in the clip. \\
        6  & Does this footage include \textless sks1\textgreater\ or \textless sks2\textgreater? & Both \textless sks1\textgreater\ and \textless sks2\textgreater\ appear in this video. \\
        7  & Can you spot \textless sks1\textgreater\ or \textless sks2\textgreater\ in this clip? & The footage contains both \textless sks1\textgreater\ and \textless sks2\textgreater. \\
        8  & Are \textless sks1\textgreater\ or \textless sks2\textgreater\ present in this video? & I can see both \textless sks1\textgreater\ and \textless sks2\textgreater\ in frame. \\
        9  & Do \textless sks1\textgreater\ or \textless sks2\textgreater\ appear in this footage? & \textless sks1\textgreater\ and \textless sks2\textgreater\ are both featured in this video. \\
        10 & Can you tell if \textless sks1\textgreater\ or \textless sks2\textgreater\ are shown here? & The recording shows both \textless sks1\textgreater\ and \textless sks2\textgreater\ present. \\
        11 & Are \textless sks1\textgreater\ or \textless sks2\textgreater\ in this video segment? & I've spotted both \textless sks1\textgreater\ and \textless sks2\textgreater\ in the footage. \\
        12 & Can you confirm \textless sks1\textgreater\ or \textless sks2\textgreater's presence? & Both \textless sks1\textgreater\ and \textless sks2\textgreater\ are captured in this clip. \\
        13 & Does this clip contain \textless sks1\textgreater\ or \textless sks2\textgreater? & The video includes both \textless sks1\textgreater\ and \textless sks2\textgreater. \\
        14 & Are \textless sks1\textgreater\ or \textless sks2\textgreater\ featured in this recording? & I can confirm the presence of both \textless sks1\textgreater\ and \textless sks2\textgreater. \\
        15 & Can you find \textless sks1\textgreater\ or \textless sks2\textgreater\ in this video? & \textless sks1\textgreater\ and \textless sks2\textgreater\ are both shown in the recording. \\
        16 & Are \textless sks1\textgreater\ or \textless sks2\textgreater\ shown in any frame? & Both figures, \textless sks1\textgreater\ and \textless sks2\textgreater, are visible. \\
        17 & Does this video show \textless sks1\textgreater\ or \textless sks2\textgreater? & I've found both \textless sks1\textgreater\ and \textless sks2\textgreater\ in the video. \\
        18 & Are \textless sks1\textgreater\ or \textless sks2\textgreater\ visible anywhere? & The footage displays both \textless sks1\textgreater\ and \textless sks2\textgreater. \\
        19 & Can you see \textless sks1\textgreater\ or \textless sks2\textgreater? & Both \textless sks1\textgreater\ and \textless sks2\textgreater\ are identifiable here. \\
        20 & Are \textless sks1\textgreater\ or \textless sks2\textgreater\ in this video? & I can recognize both \textless sks1\textgreater\ and \textless sks2\textgreater. \\
        21 & Can you recognize \textless sks1\textgreater\ or \textless sks2\textgreater? & \textless sks1\textgreater\ and \textless sks2\textgreater\ both appear in this recording. \\
        22 & Do \textless sks1\textgreater\ or \textless sks2\textgreater\ appear at all? & The video features both \textless sks1\textgreater\ and \textless sks2\textgreater. \\
        23 & Are \textless sks1\textgreater\ or \textless sks2\textgreater\ recorded here? & Both \textless sks1\textgreater\ and \textless sks2\textgreater\ are clearly visible. \\
        24 & Can you identify \textless sks1\textgreater\ or \textless sks2\textgreater? & I've detected the presence of both \textless sks1\textgreater\ and \textless sks2\textgreater. \\
        25 & Are \textless sks1\textgreater\ or \textless sks2\textgreater\ present? & The clip shows both \textless sks1\textgreater\ and \textless sks2\textgreater. \\
        \bottomrule
    \end{tabular}
     \caption{ Example of positive existence verification question answering (Two entities).}
    \label{tab:two_entities_yes}
\end{table*}

% --- Table 4: Two Entities (1 Yes, 2 No & Both No) ---

\begin{table*}[t]
    \centering
    \captionsetup{justification=centering}
    \renewcommand{\arraystretch}{1.4}
    \small
    \begin{tabular}{p{1.5cm} p{6.5cm} p{6.5cm}}  
        \toprule
        \textbf{Index} & \textbf{sks1 Yes, sks2 No} & \textbf{Both No} \\
        \midrule
        1  & Is there any trace of \textless sks1\textgreater\ or \textless sks2\textgreater\ in this footage? 
           & I can confirm that \textless sks1\textgreater\ appears, but \textless sks2\textgreater\ is not present. \newline Neither \textless sks1\textgreater\ nor \textless sks2\textgreater\ appear in this video. \\
        2  & Can you detect \textless sks1\textgreater\ or \textless sks2\textgreater\ in this video clip? 
           & The video shows \textless sks1\textgreater, though there's no sign of \textless sks2\textgreater. \newline I cannot detect either \textless sks1\textgreater\ or \textless sks2\textgreater. \\
        3  & Do \textless sks1\textgreater\ or \textless sks2\textgreater\ show up anywhere in this recording? 
           & \textless sks1\textgreater\ is visible, but \textless sks2\textgreater\ is absent. \newline The video contains neither \textless sks1\textgreater\ nor \textless sks2\textgreater. \\
        4  & Are \textless sks1\textgreater\ or \textless sks2\textgreater\ visible in this video? 
           & I've detected \textless sks1\textgreater, while \textless sks2\textgreater\ does not appear. \newline Both \textless sks1\textgreater\ and \textless sks2\textgreater\ are absent. \\
        5  & Could you verify if \textless sks1\textgreater\ or \textless sks2\textgreater\ are here? 
           & The video contains \textless sks1\textgreater, but \textless sks2\textgreater\ is not shown. \newline There is no sign of either \textless sks1\textgreater\ or \textless sks2\textgreater. \\
        6  & Does this footage include \textless sks1\textgreater\ or \textless sks2\textgreater? 
           & \textless sks1\textgreater\ is present, however \textless sks2\textgreater\ is not in this clip. \newline Neither \textless sks1\textgreater\ nor \textless sks2\textgreater\ are shown. \\
        7  & Can you spot \textless sks1\textgreater\ or \textless sks2\textgreater\ in this clip? 
           & I can see \textless sks1\textgreater, but there's no trace of \textless sks2\textgreater. \newline I confirm both \textless sks1\textgreater\ and \textless sks2\textgreater\ are not present. \\
        8  & Are \textless sks1\textgreater\ or \textless sks2\textgreater\ present in this video? 
           & The footage includes \textless sks1\textgreater, though \textless sks2\textgreater\ is not visible. \newline The footage does not include \textless sks1\textgreater\ or \textless sks2\textgreater. \\
        9  & Do \textless sks1\textgreater\ or \textless sks2\textgreater\ appear in this footage? 
           & \textless sks1\textgreater\ appears, but \textless sks2\textgreater\ is not featured. \newline There's no evidence of either \textless sks1\textgreater\ or \textless sks2\textgreater. \\
        10 & Can you tell if \textless sks1\textgreater\ or \textless sks2\textgreater\ are shown here? 
           & I've spotted \textless sks1\textgreater, while \textless sks2\textgreater\ is nowhere to be seen. \newline Neither \textless sks1\textgreater\ nor \textless sks2\textgreater\ are visible. \\
        11 & Are \textless sks1\textgreater\ or \textless sks2\textgreater\ in this video segment? 
           & \textless sks1\textgreater\ is clearly visible, but \textless sks2\textgreater\ is not. \newline I've checked, both \textless sks1\textgreater\ and \textless sks2\textgreater\ are absent. \\
        12 & Can you confirm \textless sks1\textgreater\ or \textless sks2\textgreater's presence? 
           & The recording shows \textless sks1\textgreater, though \textless sks2\textgreater\ is absent. \newline This video shows neither \textless sks1\textgreater\ nor \textless sks2\textgreater. \\
        13 & Does this clip contain \textless sks1\textgreater\ or \textless sks2\textgreater? 
           & I can identify \textless sks1\textgreater, but \textless sks2\textgreater\ doesn't appear. \newline I see no sign of \textless sks1\textgreater\ or \textless sks2\textgreater. \\
        14 & Are \textless sks1\textgreater\ or \textless sks2\textgreater\ featured in this recording? 
           & \textless sks1\textgreater\ is present, while \textless sks2\textgreater\ is not. \newline Both \textless sks1\textgreater\ and \textless sks2\textgreater\ are not in the recording. \\
        15 & Can you find \textless sks1\textgreater\ or \textless sks2\textgreater\ in this video? 
           & The clip features \textless sks1\textgreater, but there's no sign of \textless sks2\textgreater. \newline The video does not contain \textless sks1\textgreater\ or \textless sks2\textgreater. \\
        \bottomrule
    \end{tabular}
    \caption{ Example of mixed existence verification question answering (Two entities).}
    \label{tab:two_entities_mixed}
\end{table*}

\begin{table*}[!t]
    \centering
    \captionsetup{justification=centering}
    \renewcommand{\arraystretch}{1.4}
    
    \small
    \begin{tabular}{p{1.5cm} p{10cm}}  % 只有两列，第二列宽度增加
        \toprule
        \textbf{Index} & \textbf{Question} \\
        \midrule
        1  & What activity is \textless sks\textgreater\ engaged in during this video? \\
        2  & Could you describe what \textless sks\textgreater\ is doing in this footage? \\
        3  & What specific actions can you observe \textless sks\textgreater\ performing in this recording? \\
        4  & What movements or actions does \textless sks\textgreater\ perform here? \\
        5  & Can you describe \textless sks\textgreater's behavior in this sequence? \\
        6  & What is \textless sks\textgreater\ wearing in this video? \\
        7  & Could you describe \textless sks\textgreater's outfit in this footage? \\
        8  & What color and style of clothing is \textless sks\textgreater\ dressed in? \\
        9  & How would you describe \textless sks\textgreater's appearance and attire? \\
        10 & What notable features can you see in \textless sks\textgreater's clothing? \\
        11 & Where is \textless sks\textgreater\ positioned in this video? \\
        12 & What color and style of clothing is \textless sks\textgreater\ dressed in? \\
        13 & Can you describe \textless sks\textgreater's location relative to others? \\
        14 & Which part of the scene does \textless sks\textgreater\ appear in? \\
        15 & How does \textless sks\textgreater's position change throughout the video? \\
        16 & Where can \textless sks\textgreater\ be found in this footage? \\
        \bottomrule
    \end{tabular}
    \caption{ Example of negative existence verification question answering (Single entity).}
    \label{tab:single 3question}
\end{table*}

% \clearpage
% \clearpage
\section{Example of queries prompt for GPT or Internvideo}

\begin{figure*}[h]
    \vspace*{\fill} % 填充上方空白
    \centering

    % 第一张图
    \makebox[\textwidth]{\includegraphics[width=0.7\textwidth]{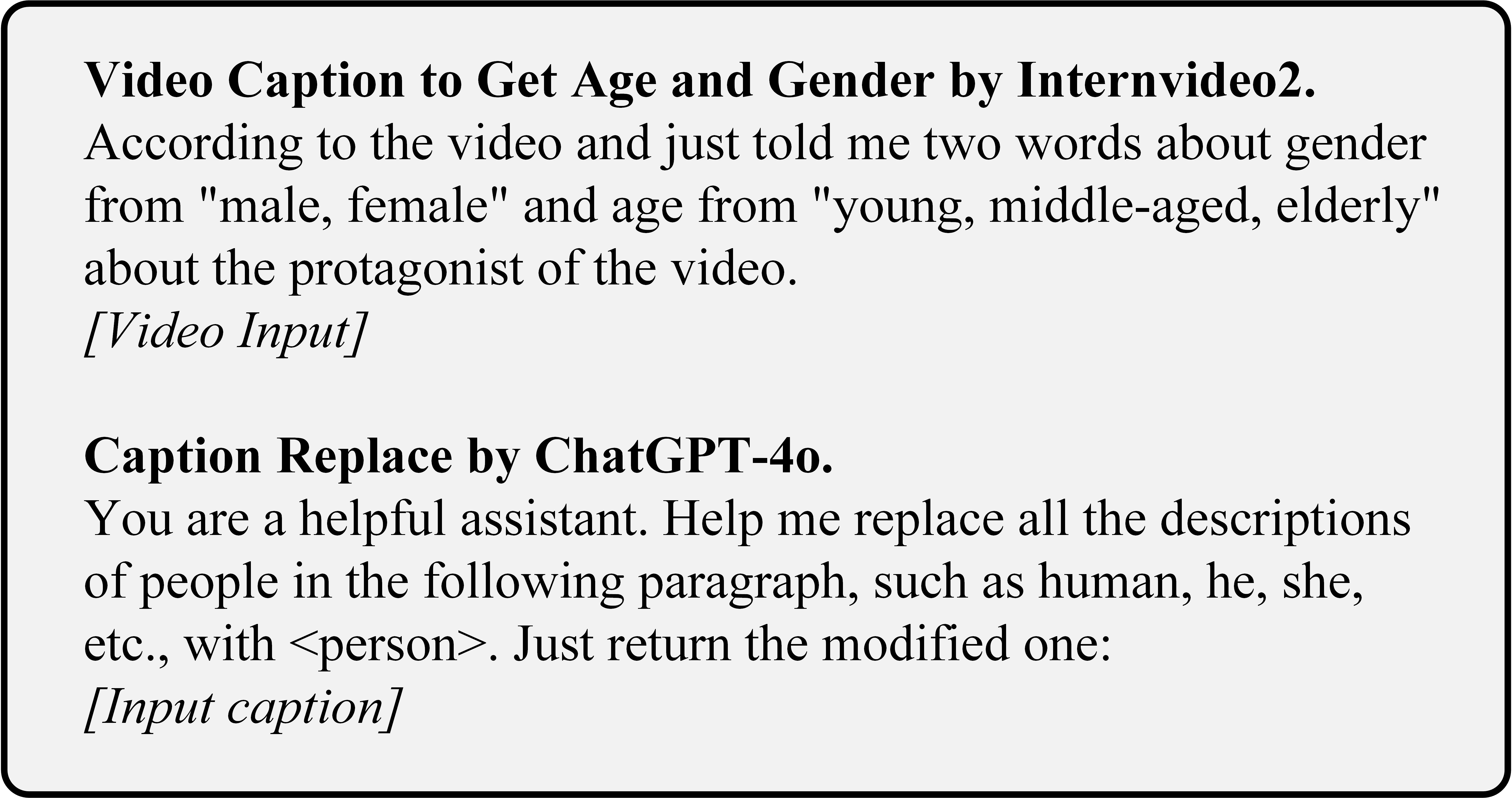}} % 强制居中
    \caption{Prompt for Internvideo and GPT query} % 限制caption宽度
    \label{fig:photo1}

    \vspace{0.5cm} % 调整两张图之间的垂直间距

    % 第二张图
    \makebox[\textwidth]{\includegraphics[width=0.7\textwidth]{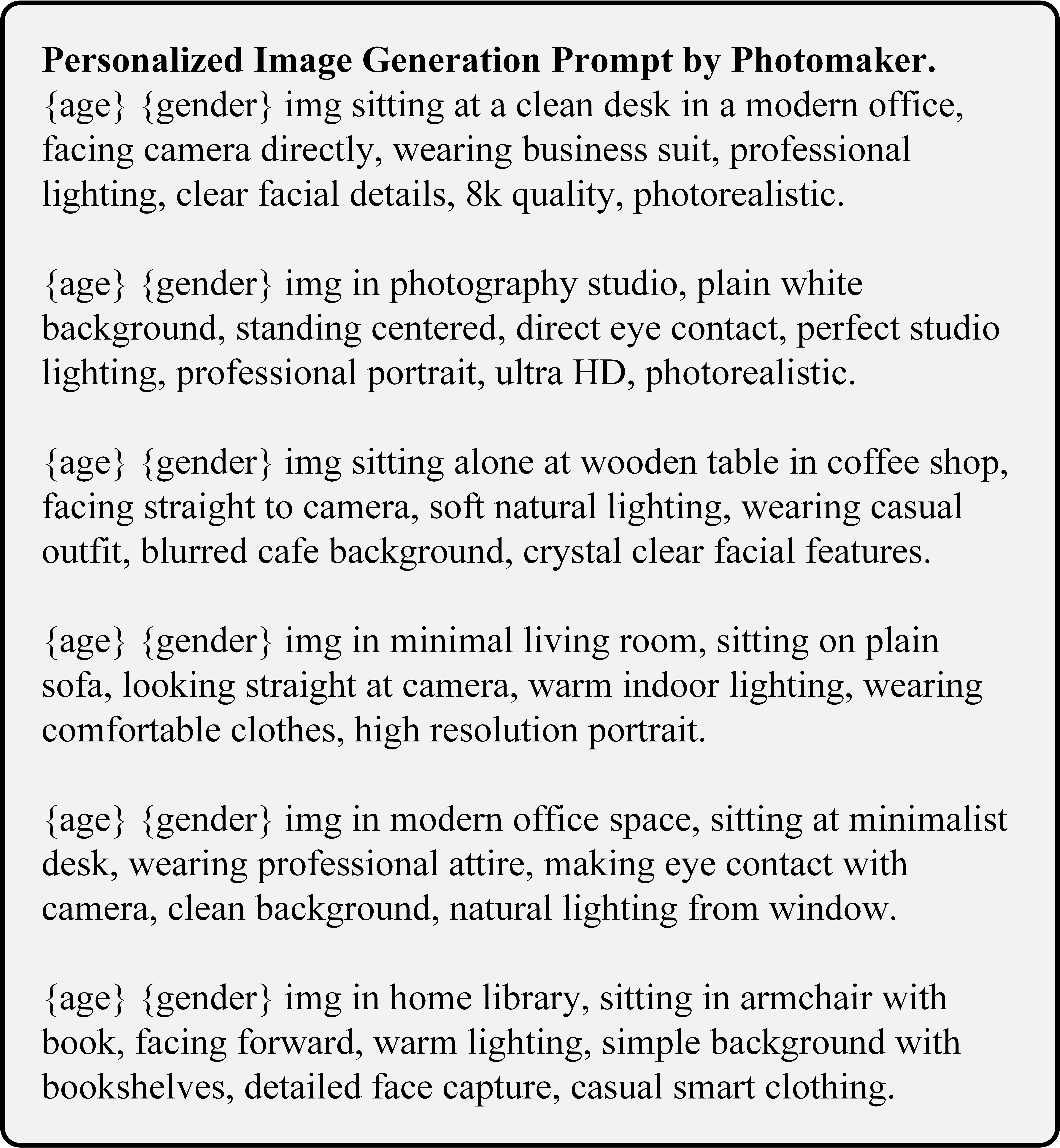}} % 强制居中
    \caption{Prompt for Photomaker synthetic the ID consist photo} % 限制caption宽度
    \label{fig:photo2}

    \vspace*{\fill} % 填充下方空白
\end{figure*}

As shown in \cref{fig:photo1}, this prompt will be used to get the age and gender of the character in the video by the video understanding model, and the age and gender information will be used in the following steps. And for the general answer from the video, understanding model will use the second one prompt to query ChatGPT to capture the general description like the human, he, she, and this will be replaced by the specific \textless sks\textgreater and then to construct our personalized QA pairs.

As shown in \cref{fig:photo2}, leveraging the character's age and gender attributes extracted in the previous stage, we enrich these core identity markers with detailed descriptors to create comprehensive prompts for (image+text)-to-image models. This approach significantly enhances identity consistency across generated images, maintaining coherent character representation while accommodating diverse poses and contexts. The resulting outputs demonstrate superior identity preservation compared to methods lacking such demographic anchoring.

\section{Presentation of 25 characters}

\begin{figure*}[htb]  % !t 表示尽量将图片放在页面顶部
    \centering
    \includegraphics[width=.8\textwidth,height=1\textwidth]{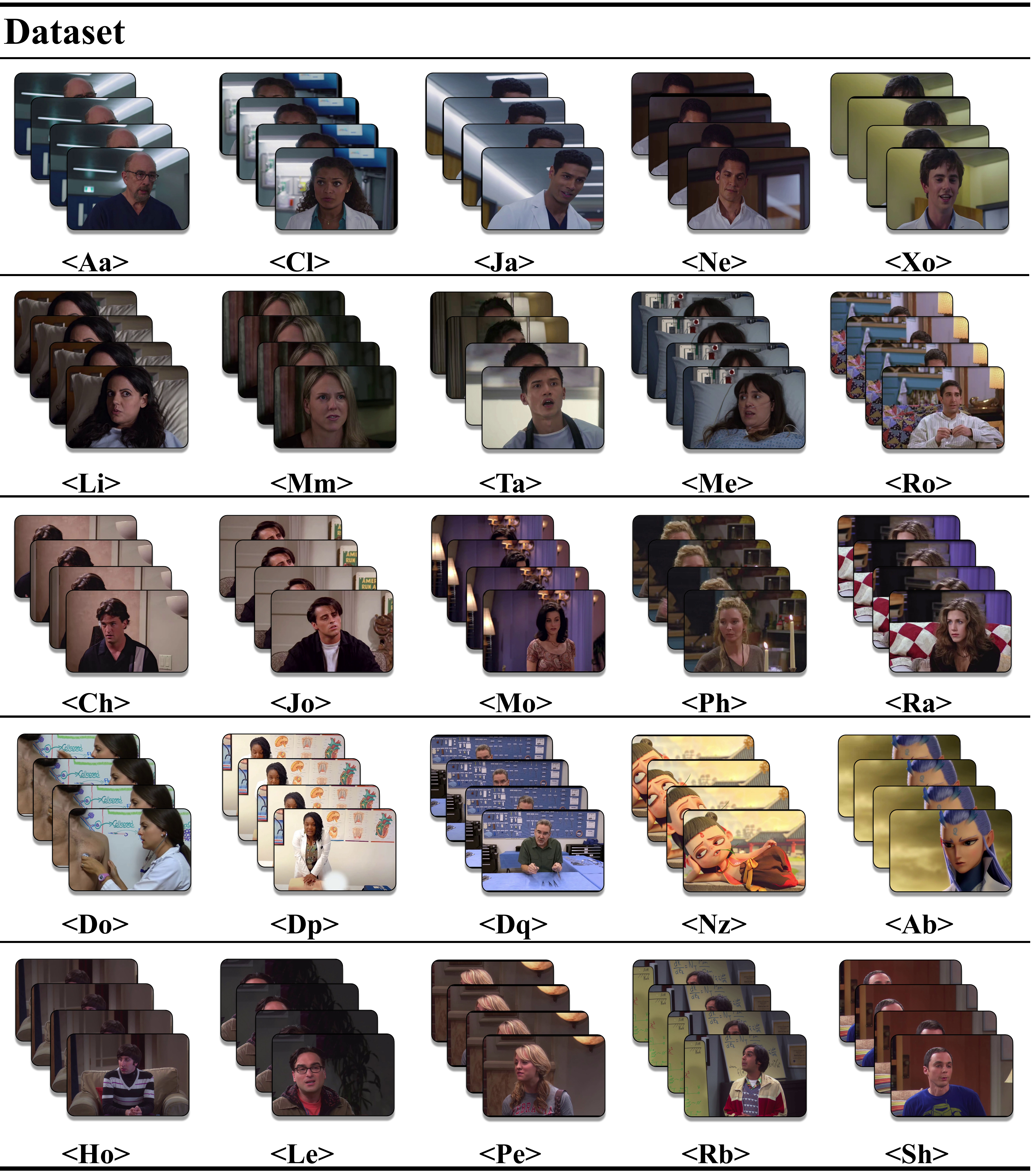}
    \caption{Display of all dataset.}
    \label{dataset}
    % \vspace{-0.2cm}
\end{figure*}

As shown in \cref{dataset}, our dataset is composed of Friends(6), Good Doctor(5), Ne Zha(2), doctor(3), patient(3), Big Bang(6)) 25 characters.

